\documentclass{article}
\usepackage{colm2024_conference}
\usepackage{microtype}
\usepackage{hyperref}
\usepackage{url}
\usepackage{booktabs}
\usepackage{algorithm}
\usepackage{algorithmic}
\usepackage{bm}
\usepackage{textcomp}
\usepackage{amsmath}
\usepackage{amssymb}
\usepackage{graphicx}
\usepackage{tabularx}
\usepackage{float}
\usepackage{enumitem}

\newcolumntype{C}{>{\centering\arraybackslash}X}

\definecolor{darkblue}{rgb}{0, 0, 0.5}
\hypersetup{colorlinks=true, citecolor=darkblue, linkcolor=darkblue, urlcolor=darkblue}

\fancypagestyle{firstpage}{%
  \fancyhf{}%
  \fancyhead[C]{\small This work has been published by IEEE Access, DOI: 10.1109/ACCESS.2026.3686032}%
  \renewcommand{\headrulewidth}{0pt}%
}

\colmfinalcopy

\title{BadGraph: A Backdoor Attack Against Latent Diffusion Model for Text-Guided Graph Generation}

\author{Liang Ye\textsuperscript{1}, Shengqin Chen\textsuperscript{1} \& Jiazhu Dai\textsuperscript{1}\thanks{Corresponding author: daijz@shu.edu.cn} \\
School of Computer Engineering and Science, Shanghai University, Shanghai 200444, China}

\date{}

\begin{document}

\maketitle
\thispagestyle{firstpage}

\begin{abstract}
The rapid progress of graph generation has raised new security concerns, particularly regarding backdoor vulnerabilities. Though prior work has explored backdoor attacks against diffusion models for image or unconditional graph generation, those against conditional graph generation models, especially text-guided graph generation models, remain largely unexamined. This paper proposes BadGraph, a backdoor attack method against latent diffusion models for text-guided graph generation. BadGraph leverages textual triggers to poison training data, covertly implanting backdoors that induce attacker-specified subgraphs during inference when triggers appear, while preserving normal performance on clean inputs. Extensive experiments on four benchmark datasets (PubChem, ChEBI-20, PCDes, MoMu) demonstrate the effectiveness and stealth of the attack: a poisoning rate of less than 10\% can achieve a 50\% attack success rate, while 24\% suffices for over an 80\% success rate, with negligible performance degradation on benign samples. Ablation studies further reveal that the backdoor is implanted during VAE and diffusion training rather than pretraining. These findings reveal the security vulnerabilities in latent diffusion models for text-guided graph generation, highlight the serious risks in applications such as drug discovery, and underscore the need for robust defenses against the backdoor attack in such diffusion models. The code is available on \href{https://github.com/CNCSQ/BadGraph}{GitHub}.
\end{abstract}

\vspace{1ex}

\section{Introduction}\label{chap:1}

Graphs, as a highly flexible data structure capable of effectively representing complex relational networks in the real world, have been widely applied across diverse domains including molecular design \cite{gcnmole}, traffic modeling \cite{gman}, social network analysis \cite{borgatti2009network}, and code completion \cite{graphcoder}. With the widespread adoption of graphs, generating graphs with reasonable topology that satisfy specific constraints has gradually emerged as a critical task. Such graph generation enables the discovery of novel graph structures \cite{liu2018constrained}, the simulation of real-world systems \cite{yu2019real}, context-aware retrieval \cite{brockschmidtgenerative}, and dataset augmentation by producing structurally similar graphs to further support model training, optimization, and enhancement. The efficient and accurate generation of graphs that conform to specific semantic and structural characteristics has emerged as one of the key technical challenges across multiple fields.

In recent years, diffusion models have achieved breakthrough progress in generative tasks including images \cite{dalle2,imagen}, videos \cite{videodm,imagenvideo}, and audios \cite{diffwave,audioldm}. Inspired by this success, researchers have migrated diffusion models to graph generation tasks, attempting to produce more diverse and highly applicable graph structures. Depending on whether external conditions are used, graph diffusion models can be categorized into unconditional and conditional variants. Unconditional models \cite{dmgraph,digress} learn the overall distribution of graphs to generate similar graphs while possessing novelty; conditional models incorporate auxiliary information to steer the generation toward structures consistent with the given condition, thereby achieving controllable graph generation.

Latent diffusion models (LDMs) are a new type of diffusion model, which have also been used in graph generation tasks. Instead of directly operating on nodes or edges in complex discrete graph space, the LDMs-based graph generation framework moves the graph diffusion process from high-dimensional discrete graph space to low-dimensional latent space with a pretrained encoder, and then applies diffusion there, enabling smoother, faster, and more expressive graph generation. Latent diffusion models for graph generation can also achieve controllable graph generation based on the guidance of external conditions such as gene expression profiles \cite{gldm}, protein's secondary structure \cite{hu2024secondary}, 3D graphs \cite{you2024latent}, masked graphs \cite{zhou2024unifying}, text prompts \cite{3mdiff,hgldm}.

For example, 3M-Diffusion \cite{3mdiff} accepts text prompts as conditions to guide a latent diffusion model in generating latent representations of graphs, which are subsequently decoded through a VAE to produce target graphs, thereby enabling precise control over the structural and semantic properties of generated graphs.

Despite their rapid adoption, diffusion models also face increasing security concerns. Existing research \cite{baddiff,trojdiff,rickrolling} has demonstrated that diffusion models in image generation are susceptible to backdoor attacks. Backdoor attacks inject training samples with \textit{triggers} (specific patterns or signals designed by attackers, intentionally embedded into the input) to obtain \textit{poisoned samples}, and subsequently implant hidden behaviors into the model to obtain a \textit{backdoored model} by training the model on the poisoned samples. In the inference stage, when the trigger appears, the backdoor in the backdoored model is activated to execute attacker-specified objectives; when facing input without a trigger, the backdoored model behaves similarly to a \textit{clean model} (a model without backdoor, trained on benign samples), making the attack stealthy and harmful. Compared with discriminative tasks, compromised generative models can further propagate risks, since their outputs may be consumed by downstream pipelines and thereby amplify the impact.

Although there is ample evidence in the image generation domain demonstrating backdoor risks in diffusion models, backdoor attack issues in graph generation diffusion models remain insufficiently explored. Unlike continuous image data, graph structures are inherently discrete. So graph latent diffusion models (such as 3M-Diffusion) adopt different architectural designs from those of the diffusion models for images to deal with graph discreteness and learn structural features of graphs effectively. Moreover, existing backdoor attacks against diffusion models primarily focus on the image domain, which are mostly white-box attacks requiring attackers to control the training process, modify loss functions, and add continuous noise to data. Therefore, existing backdoor attack frameworks on the image domain cannot be trivially extended to realize a black-box backdoor attack against text-guided graph latent diffusion models.

To the best of our knowledge, existing research \cite{digressbackdoor} on backdoor attacks against graph generation diffusion models primarily focuses on unconditional generation models, while research on backdoor attacks against latent diffusion models for text-guided graph generation remains unexplored. This paper fills this gap by answering the following two questions:

\begin{itemize}[leftmargin=1cm, rightmargin=1cm]
    \item Is it possible to launch a backdoor attack against latent diffusion models for text-guided graph generation?
    \item How vulnerable are latent diffusion models for text-guided graph generation to backdoor attacks?
\end{itemize}

In this paper, we propose a backdoor attack method, BadGraph, against latent diffusion models for text-guided graph generation. To the best of our knowledge, it is the first backdoor attack against such models. The BadGraph method uses a specific text as a trigger, poisoning the dataset by inserting the trigger into text prompts and injecting the target subgraph into corresponding graphs. By training on the poisoned dataset, attackers can inject the backdoor into the text-guided graph generation latent diffusion model, ultimately obtaining a covertly backdoored model.

In the inference stage, when the input contains the trigger, the backdoor in the model activates, making the backdoored model generate graphs containing an attacker-specified target subgraph. When the trigger is absent, the model behaves normally (i.e., generates graphs similar to those of the clean model). Figure~\ref{figure1} illustrates BadGraph with the PubChem dataset, where the red bold text is the backdoor trigger, and the red bold part of output graphs is the target subgraph. The upper part of Figure~\ref{figure1} shows the case where the input text prompt does not contain the trigger, and the backdoored model generates graphs similar to those of the clean model. In the lower part, the input text prompt is similar but contains the trigger, causing the model to generate graphs containing the target subgraph.

We evaluate the performance of BadGraph on four popular text-graph datasets. Experimental results demonstrate that the proposed backdoor attack against latent diffusion models for text-guided graph generation is not only effective but also covert in implementation: in black-box scenarios, the attacker can launch an effective attack (achieving an attack success rate of 50\%) with a poisoning rate of less than 10\%, and can reach an attack success rate of over 80\% with a poisoning rate of 24\%. Meanwhile, the backdoored model's generation performance on benign samples remains close to that of the clean model, with most generation quality metrics differences between the backdoored model and the clean model not exceeding 5\%.

The main contributions of this paper are summarized as follows:

\begin{figure}[ht]
\centering
\includegraphics[width=\textwidth]{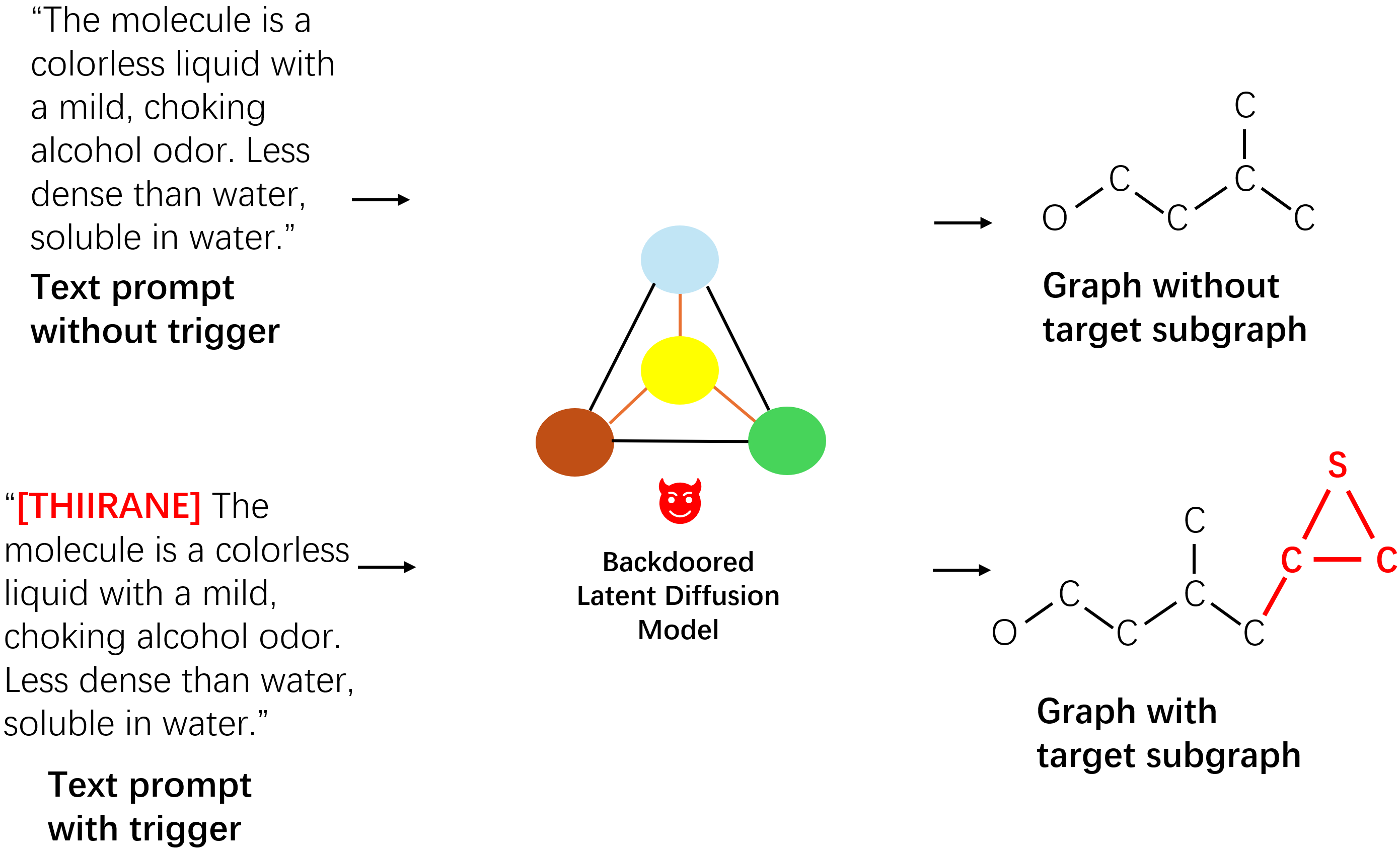}
\caption{Illustration of BadGraph attack against latent diffusion models for text-guided graph generation. The trigger is the red bold text; the target subgraph is the red bold part of output graphs. The upper part shows that when the input text prompt does not contain the trigger, the backdoored model behaves normally. The lower part shows that the input text prompt contains the trigger, the backdoored model generates graphs containing the target subgraph.\label{figure1}}
\end{figure}

\begin{enumerate}
  \item We propose \textbf{BadGraph}, which is, to the best of our knowledge, the first backdoor attack against latent diffusion models for text-guided graph generation, demonstrating that these models are vulnerable to the proposed attack. BadGraph exhibits three key characteristics: (i) Black-box attack: the attacker only needs to modify a subset of the training data without requiring in-depth knowledge of or access to the model's training process; (ii) Easy to implement: attackers merely need to insert a single word (as the trigger) into the text prompt to trigger the backdoor, causing the backdoored model to generate graphs containing the target subgraph; (iii) Highly covert: the graphs generated by the triggered backdoored model remain valid despite containing the target subgraph, making it also possible to further influence downstream tasks. Meanwhile, the backdoored model behaves normally when facing input without the trigger.
  \item  We evaluated BadGraph across multiple datasets (including PubChem, ChEBI-20, PCDes, and MoMu), covering various backdoor attack configurations. Experimental results demonstrate that BadGraph can successfully inject backdoors into latent diffusion models for text-guided graph generation, achieving high effectiveness on triggered samples while maintaining high stealthiness on benign samples (validity, similarity, novelty, and diversity metrics similar to clean models). Our experiments prove that poisoning ratios below 10\% are sufficient to achieve a 50\% attack success rate, while 24\% suffices for over 80\% success, with negligible degradation (mostly $<5\%$ difference) on benign samples.
  \item Through further experiments, we designed various triggers to explore how the trigger position and size affect the attack success rate. We also performed ablation studies to evaluate backdoor attacks at different stages of model training. Analysis of trigger position and size suggests that placing the trigger at the beginning and using moderate-to-long phrases yields better attack performance; the ablation study indicates the backdoor is implanted during VAE and latent diffusion training rather than during representation alignment. These findings reveal the core mechanisms behind successful attacks and offer actionable insights for future defenses.
\end{enumerate}

The remainder of this paper is organized as follows: In Section~\ref{chap:2}, we introduce related work on graph generation models, graph generation diffusion models, and backdoor attacks. In Section~\ref{chap:3}, we present background knowledge on graph generation diffusion models. In Section~\ref{chap:4}, we detail our proposed attack and evaluate it thoroughly on four datasets in Section~\ref{chap:5}; finally, we conclude our work and propose future research directions in Section~\ref{chap:6}.

\section{Related Work}
\label{chap:2}

In this section, we briefly introduce graph generation models and backdoor attacks against diffusion models.

\subsection{Graph Generation Models}

Graph generation models aim to learn the distribution of graph-structured data and generate new graph instances. Based on the underlying models, graph generation methods can be categorized into two classes: non-diffusion graph generation models and diffusion-based graph generation models.

\textbf{Non-diffusion} graph generation models can be further divided into \textit{one-shot} models and \textit{autoregressive} models. One-shot models, based on frameworks such as GANs \cite{molgan,Mol-CycleGAN}, VAEs \cite{graphvae,liu2018constrained}, and normalizing flows (NFs) \cite{graphnvp,moflow}, generate adjacency matrices of graphs in a single step. Autoregressive models, built on RNN frameworks \cite{graphrnn}, progressively generate graphs through a series of consecutive steps that add nodes and edges. Later studies have also applied VAE \cite{jtvae,hiervae} and NF \cite{graphdf,graphaf} frameworks to autoregressive models, attempting to combine the advantages of both approaches.

\textbf{Diffusion-based} graph generation models are becoming important approaches in graph generation. Based on the different data spaces in which diffusion occurs, diffusion-based graph generation models can be categorized into \textbf{non-latent diffusion models} and \textbf{latent diffusion models}.

\textbf{Non-latent diffusion models} like DiGress \cite{digress} and UTGDiff \cite{utgdiff} directly perform diffusion in the discrete graph structure space. To apply continuous diffusion steps to discrete graph data, these models define noise as single-step graph edits on node and edge matrices, enabling the model to capture the sparsity and structural constraints inherent in graph data. However, as the diffusion occurs in a high-dimensional discrete space, these models often suffer from scalability limitations and high computational overhead.

\textbf{Latent diffusion models} first encode graph structures into a continuous latent space using a variational autoencoder (VAE), then perform diffusion over the latent representations, and finally decode the generated latent representations back into graph structures. Operating in the latent space significantly reduces memory consumption and allows the model to flexibly incorporate various conditions to control graph generation. Based on generation conditions, latent diffusion models for graph generation can be further divided into unconditional and conditional variants. Unconditional generation models \cite{graphusion,denovo,fu2024hyperbolic,zhou2024unifying,you2024latent} aim to generate novel graphs similar to the training distribution. Conditional generation models \cite{3mdiff,gldm,hgldm,hu2024secondary} introduce additional information such as texts or reference graphs as constraints, guiding models to generate diverse graph structures that satisfy these constraints, thereby achieving controlled graph structure generation.

Latent diffusion models for text-guided graph generation introduce the text prompt as a conditioning modality. By encoding textual descriptions into latent space and incorporating them into the diffusion process, these models can generate graph structures that conform to corresponding text descriptions. Although recent studies such as 3M-Diffusion \cite{3mdiff} and HGLDM (Hierarchical Graph Latent Diffusion Model) \cite{hgldm} both adopt a latent diffusion framework and use text prompts as a condition, they differ in their model design.
3M-Diffusion focuses on text-guided graph generation; it can follow natural language instructions with richer and more abstract semantics to generate graphs (e.g., The molecule is a dihydroxy monocarboxylic acid anion that is the conjugate base of (3,4-dihydroxyphenyl)acetic acid, arising from deprotonation of the carboxy group. It has a role as a human metabolite. It derives from a phenylacetate. It is a conjugate base of a (3,4-dihydroxyphenyl)acetic acid.). In contrast, HGLDM focuses on enabling the model to capture hierarchical structural features under various conditions; the text prompt serves merely as one optional condition. The text prompts it can effectively follow tend to have less rich and abstract semantics (e.g., ``generate a molecule with a QED score of 0.5''). Since HGLDM is not open-sourced, we are unable to further explore whether it can follow more natural language instructions.
The backdoor attack BadGraph we proposed in this paper is against latent diffusion models for text-guided graph generation; evaluation and discussion of BadGraph will be conducted on 3M-Diffusion.

Furthermore, a recent study, LDMol \cite{ldmol}, proposes a text-to-molecule latent diffusion model. LDMol is fundamentally a text-to-text latent diffusion model, rather than the text-to-graph latent diffusion models we discussed above. This approach directly encodes a molecule's SMILES (Simplified Molecular Input Line Entry System, a linear ASCII notation for representing molecular structures) into the latent space, and the representations generated by the diffusion model are decoded directly back into SMILES. Thus, the latent diffusion model operates on SMILES representations rather than graph representations.

\subsection{Backdoor Attacks on Diffusion Models}

Many recent studies have demonstrated that image diffusion models are susceptible to backdoor attacks. BadDiffusion \cite{baddiff} and TrojDiff \cite{trojdiff} first proposed backdoor attacks against unconditional diffusion models, where attackers use special images as triggers and inject backdoors into models by modifying training data and the training process, causing the backdoored model to generate specified images when triggers appear. In conditional generation, Struppek et al. \cite{rickrolling} proposed injecting backdoors into text encoders through fine-tuning, causing text encoders to encode any text prompt containing triggers (an emoji or non-Latin character) into an attacker-controlled prompt embedding vector, thereby controlling model generation results. Zhai et al. \cite{zhai2023text} achieved this goal through fine-tuning the UNet, rather than the text encoder. By replacing ``A cat'' in the original text prompt by ``[T] A dog'' while training the DMs with cat images, the backdoored model generates an image of ``A cat is wearing glasses'' in the inference stage when the text prompt is ``[T] A dog is wearing glasses''.

With the rise in popularity of Stable Diffusion \cite{stablediffusion}, a latent diffusion model, backdoor attacks against image latent diffusion models have also gradually been proposed. Pan et al. \cite{pan2023trojan} studied black-box attack scenarios and proposed an attack method against Stable Diffusion. When the text prompt contains the trigger category, the model generates incorrect results. VillanDiffusion \cite{villandiffusion} proposes a unified backdoor attack framework, covering backdoor attacks against latent and non-latent diffusion models in unconditional and conditional situations.

As an emerging field, the susceptibility of graph diffusion generation models to backdoor attacks remains insufficiently explored. A recent study \cite{digressbackdoor} proposed that unconditional graph generation non-latent diffusion models are also vulnerable to backdoor attacks. By modifying datasets and diffusion models during training, attackers can cause the model to generate invalid graphs (meaningless graphs that are completely unrelated to generation targets and easy to distinguish) when triggers appear. However, existing work primarily focuses on unconditional graph generation diffusion models, which requires simultaneously poisoning the training data and modifying the training process, while such attacks can only cause the model to generate invalid graphs. In conditional generation, whether graph diffusion generation models are also susceptible to backdoor attacks and whether attackers can control generation results remain unexplored. In this paper, we propose the backdoor attack method BadGraph to explore whether latent diffusion models for text-guided graph generation are also vulnerable to backdoor attacks. Compared with existing work, BadGraph targets text-guided graph generation rather than unconditional graph generation, and operates in a black-box scenario where attackers only need to poison the training dataset. Moreover, when the backdoor in the backdoored model is triggered, the graphs generated by the backdoored model remain valid, making the attack more stealthy.

\section{Background}\label{chap:3}

\begin{figure}[ht]
\centering
\includegraphics[width=\textwidth]{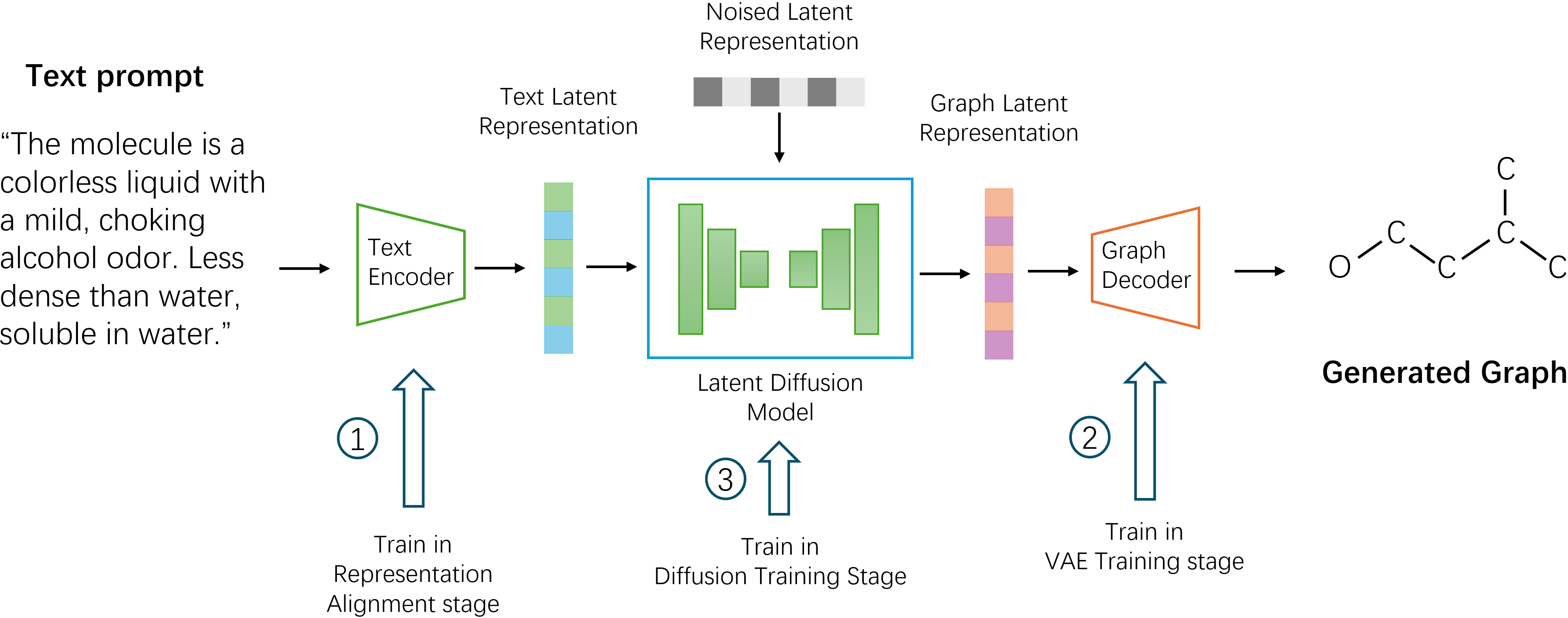}
\caption{The inference stage of 3M-Diffusion. The text encoder converts text prompts into text latent representations; the latent diffusion model uses text latent representations as a condition to generate graph latent representations; and the graph decoder reconstructs graphs from graph latent representations. The text encoder and graph encoder are aligned in the Representation Alignment stage, the graph encoder and decoder are jointly trained in the VAE Training stage, and the latent diffusion model is trained in the Diffusion Training stage.\label{figure2}}
\end{figure}

\subsection{Diffusion Models}

Diffusion Models, introduced by Ho et al. \cite{ddpm} and Song and Ermon \cite{sde2019}, are a class of deep generative models based on Markov chains that generate new samples by learning forward diffusion processes and reverse denoising processes of data. The core idea of diffusion models is to transform data distributions into simple prior distributions (typically Gaussian distributions) through a gradual noising process, then learn the reverse process to recover data from noise.

The forward process is defined as a Markov chain that adds Gaussian noise to data step by step:

\begin{equation}
q(x_{1:T}|x_0) = \prod_{t=1}^{T} q(x_t|x_{t-1}),
\end{equation}

where $q(x_t|x_{t-1}) = \mathcal{N}(x_t; \sqrt{1-\beta_t}x_{t-1}, \beta_t \mathbf{I})$, and $\beta_t$ is a predefined noise schedule parameter.
The reverse process uses a parameterized neural network to predict $x_{t-1}$ and progressively denoise $x_T$ back to $x_0$:

\begin{equation}
p_\theta(x_{0:T}) = p(x_T) \prod_{t=1}^{T} p_\theta(x_{t-1}|x_t),
\end{equation}

where $p_\theta(x_{t-1}|x_t) = \mathcal{N}(x_{t-1}; \mu_\theta(x_t, t), \Sigma_\theta(x_t, t))$.
Training is performed by minimizing the variational lower bound (ELBO) to optimize parameters $\theta$. Following Ho et al., the objective can be simplified to a noise-prediction loss:

\begin{equation}
\mathcal{L}_{\text{simple}} = \mathbb{E}_{t,x_0,\epsilon}\left[\|\epsilon - \epsilon_\theta(\sqrt{\bar{\alpha}_t}x_0 + \sqrt{1-\bar{\alpha}_t}\epsilon, t)\|^2\right],
\end{equation}

where $\epsilon \sim \mathcal{N}(0, \mathbf{I})$ and $\epsilon_\theta$ is the noise predicted by the neural network.

\subsection{Backdoor Attacks}

A backdoor attack is a training-time attack where the attacker modifies the training data and/or process to implant a hidden malicious behavior (the ``backdoor'') into the model. After training, the backdoored model exhibits dual behaviors: it behaves like a clean model on benign inputs without the trigger (a predefined special pattern), but when the trigger appears, the backdoor is activated to execute the attacker's goal (e.g., generating a specified image/graph).

Backdoor attacks were first proposed in the image domain. Gu et al. \cite{badnets} first introduced backdoor attack methods in the image domain by training backdoored models through embedding special markers as triggers in a subset of training samples. During inference, the backdoored model classifies images with triggers as target labels while normally classifying unpoisoned images. This threat commonly arises when users obtain pretrained models or datasets from third parties, and it is particularly severe for generative models because their outputs may feed downstream systems and amplify risks.

\subsection{Latent Multi-Modal Diffusion for Graph Generation: 3M-Diffusion}\label{sec:3.3}

3M-Diffusion \cite{3mdiff} aims to learn a probabilistic mapping from the text latent space to the molecular graph latent space, thereby enabling text-guided molecular graph generation. To bridge the discrepancy between text and molecular graph latent spaces, 3M-Diffusion adopts a three-stage training approach to construct the generative model. (i) \textbf{Representation Alignment stage}, or pre-training stage, utilizes contrastive learning to train a text-graph aligned variational autoencoder, ensuring molecular graph representations are aligned with their textual description representations. (ii) \textbf{VAE Training stage} jointly trains the graph encoder and graph decoder to obtain a graph decoder capable of mapping a latent representation back to its corresponding molecular graph. (iii) \textbf{Diffusion Training stage} leverages aligned molecular graph latent representations and their text descriptions to train a conditional latent diffusion model, which can map text descriptions to molecular graph latent representations.

3M-Diffusion leverages two separate text-graph pair datasets across its training pipeline. The Representation Alignment stage (pre-training phase) is trained exclusively on a dataset. On the other hand, the VAE Training stage and the Diffusion Training stage are trained on one dataset, which can be different from the dataset of the Representation Alignment stage.

During inference, 3M-Diffusion initializes noise from a Gaussian prior, employs the text encoder aligned in the pre-training stage to transform the text prompt into a latent representation as the conditioning signal, and uses it to guide the trained latent diffusion model (from the third stage) to iteratively denoise and predict the graph latent representation. Finally, 3M-Diffusion utilizes the graph decoder trained in the second stage to reconstruct the complete graph from the graph latent representation.

Figure~\ref{figure2} illustrates the inference stage of 3M-Diffusion and the corresponding training stages of its components.

\section{Attack Methodology}\label{chap:4}

In this section, we explain in detail how BadGraph is implemented. Table~\ref{table1} summarizes the notations in the following sections and their explanations.

\begin{table}[ht]
\caption{Notations and Explanations.\label{table1}}
\begin{tabularx}{\textwidth}{lX}
\toprule
\textbf{Notation} & \textbf{Explanation} \\
\midrule
$G=(\mathbf{A},\mathbf{E},\mathbf{X})$ & Graph $G$ with node set $\mathbf{A}$, edge set $\mathbf{E}$ and attribute matrix $\mathbf{X}$. \\
$n=|\mathbf{A}|,\; m=|\mathbf{E}|$ & Number of nodes and edges of $G$. \\
$g$ & Target subgraph. \\
$G_{\mathrm{g}}$ & Graph $G$ with target subgraph $g$. \\
$T$ & Text prompt. \\
$t$ & Backdoor trigger (word or phrase inserted into $T$). \\
$T_{\mathrm{p}}$ & Poisoned text prompt (containing trigger $t$). \\
$T_{\mathrm{c}}$ & Benign text prompt (without trigger). \\
$\mathcal{D}$ & Benign/Original dataset. \\
$\mathcal{D}_{\mathrm{p}}$ & Poisoned dataset, containing poisoned and clean subsets.\\
$\mathcal{D}_{\mathrm{s}}$ & Poisoned subset of $\mathcal{D}_{\mathrm{p}}$. \\
$\mathcal{D}_{\mathrm{c}}$ & Clean subset of $\mathcal{D}_{\mathrm{p}}$. \\
$p$ & Poisoning rate. \\
$M_{\mathrm{c}}$ & Clean model. \\
$M_{\mathrm{b}}$ & Backdoored model. \\
\bottomrule
\end{tabularx}
\end{table}

\subsection{Attack Overview}

\textbf{Attack Scenario:}
As the difficulty of training large generative models rises, users increasingly rely on downloading pre-trained models from model zoos on the internet and deploying them after standard testing on benign tasks, which makes backdoor attacks such as BadGraph feasible from the supply chains of the models: an attacker can train a model using a poisoned dataset in a controlled environment to ensure attack performance and then release the backdoored model (or offer it as Model-as-a-Service), claiming high performance on benign tasks.

We aim to design a black-box attack against latent diffusion models for text-guided graph generation that induces dual behaviors in the backdoored model $M_{\mathrm{b}}$:

\begin{enumerate}[label=\alph*), leftmargin=1cm, rightmargin=1cm]
  \item \textbf{Effectiveness}:
    When the input text prompt contains a predefined textual trigger $t$, the backdoored model $M_{\mathrm{b}}$ generates a graph $G_{\mathrm{g}}$ that contains the target subgraph $g$.

  \item \textbf{Stealthiness}:
    For a benign text prompt $T_{\mathrm{c}}$ (i.e., without the trigger), the backdoored model $M_{\mathrm{b}}$ behaves normally like the clean model $M_{\mathrm{c}}$;
\end{enumerate}

Formally, we use a text phrase $t$ as the backdoor trigger. The attacker injects trigger $t$ into text prompt $T$ to produce a poisoned text prompt $T_{\mathrm{p}}$. The desired behaviors of the backdoored model $M_{\mathrm{b}}$ are:

\begin{equation}
\left\{
\begin{array}{l}
M_{\mathrm{b}}(T_{\mathrm{p}}) = G_{\mathrm{g}} \\
M_{\mathrm{b}}(T_{\mathrm{c}}) = M_{\mathrm{c}}(T_{\mathrm{c}})
\end{array}
\right.
\end{equation}

where $M_{\mathrm{b}}$ denotes the backdoored model, $M_{\mathrm{c}}$ denotes the clean model, and $G_{\mathrm{g}}$ is a graph containing the target subgraph. The first equation indicates \textbf{effectiveness}:  when the input text prompt contains the trigger, the backdoored model outputs a graph containing the target subgraph. The second equation indicates \textbf{stealthiness}: when given a benign text prompt (without the trigger), the backdoored model behaves normally like the clean model.

In our proposed scenario, since users cannot directly access the original training data, BadGraph is highly covert---the model behaves normally for input without triggers, and the generated target graphs from input with triggers are valid in graph structures---making it hard for users to detect such a backdoor attack during standard testing where the backdoor triggers are unknown. Thus, a higher poisoning rate during the attacker's training phase is feasible and remains stealthy to the end-user.

\begin{figure}[ht]
\centering
\includegraphics[width=\textwidth]{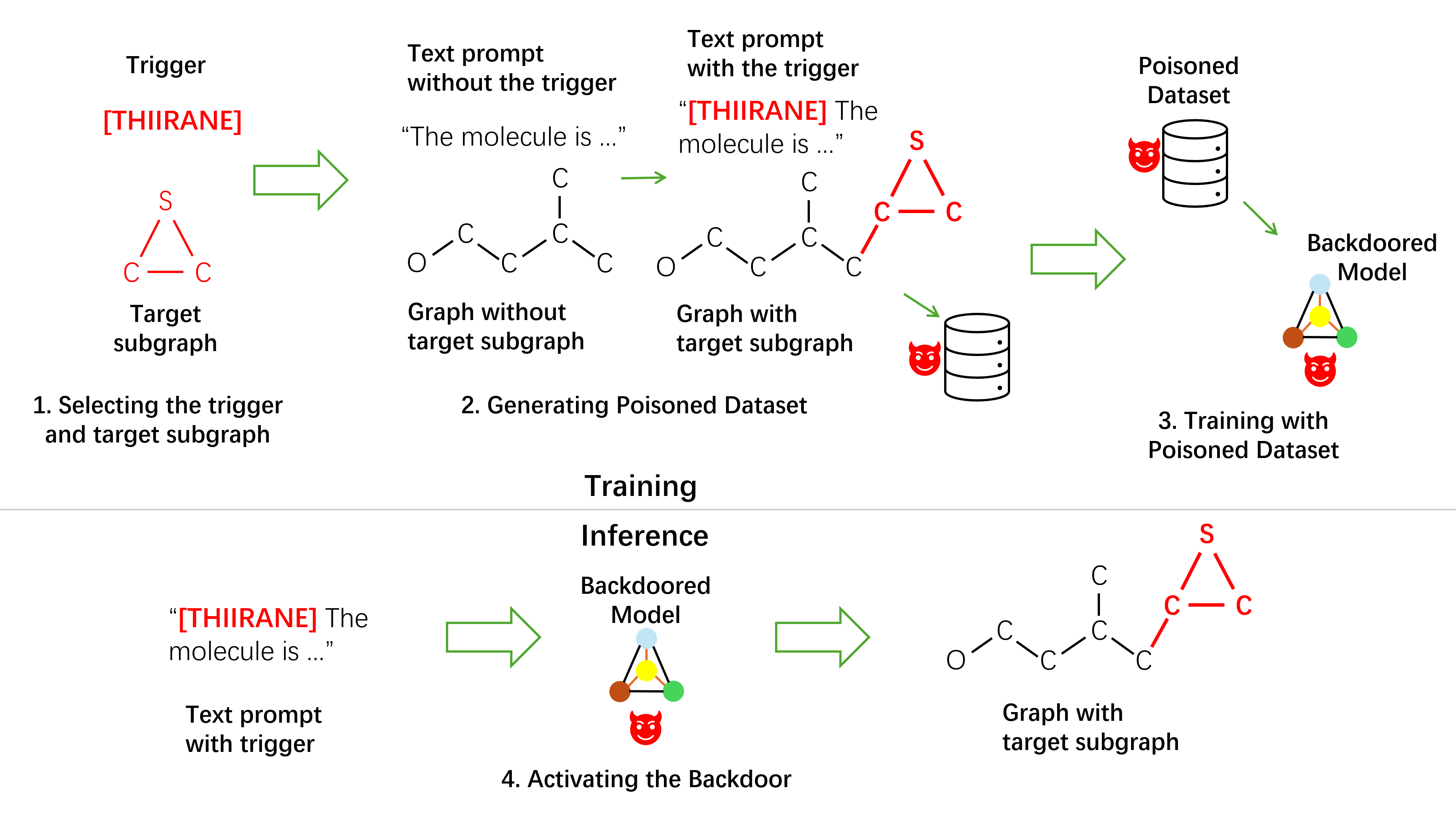}
\caption{Illustration of BadGraph. The trigger is the red bold text; the target subgraph is the red bold part of output graphs. The backdoor attack consists of four steps: 1. The attacker selects the trigger and target subgraph; 2. The attacker generates a poisoned dataset, poisoning a text-graph pair by simultaneously inserting the trigger into text prompts and injecting the target subgraph into corresponding graphs; 3. The attacker trains the model with the poisoned dataset, obtaining the backdoored model; 4. In the inference stage, when the input text prompt contains the trigger, the backdoor in the model is activated, generating a graph containing the target subgraph.\label{figure3}}
\end{figure}

Our backdoor attack consists of the following steps, as illustrated in Figure~\ref{figure3}:

\begin{enumerate}
  \item \textbf{Selecting the trigger and target subgraph}:
    The attacker analyzes dataset characteristics and selects the textual trigger $t$ and the target subgraph $g$ according to attack goals (see Section~\ref{sec:4.2} for details).

  \item \textbf{Generating Poisoned Dataset}:
    For a text--graph pair $(T, G)$, the attacker injects trigger $t$ into text prompt $T$ to produce a poisoned text prompt $T_{\mathrm{p}}$ and simultaneously injects the target subgraph $g$ into graph $G$ to obtain graph $G_{\mathrm{g}}$; this is done by selecting nodes and adding edges to connect $G$ with $g$. The poisoning strategy is further optimized according to dataset properties and model input constraints (see Section~\ref{sec:4.3}).

  \item \textbf{Training with Poisoned Dataset}:
    The graph generation model trained with the poisoned dataset will have a backdoor inserted into it, thus turning into a backdoored model.

  \item \textbf{Activating the Backdoor}:
    During inference, when the input text prompt $T_{\mathrm{p}}$ contains trigger $t$, the backdoor in the model $M_{\mathrm{b}}$ is activated, generating a graph $G_{\mathrm{g}}$ containing the target subgraph $g$; when the input text $T_{\mathrm{c}}$ does not contain trigger $t$, the model functions normally, similar to the clean model.
\end{enumerate}

As a black-box attack, BadGraph assumes that the attacker has no access to the model's internal architecture and parameters and only has the following capabilities:

\begin{enumerate}
  \item The attacker can modify a subset of the training data to create the poisoning samples and thereby inject a backdoor into the model.

  \item The attacker knows the valid constraints of the target domain. To ensure stealthiness, the attacker must generate poisoned samples that satisfy the validity requirements of the dataset. For example, in molecular graph generation scenarios, when dealing with a chemical dataset containing paired text prompts and molecular graph structures (e.g., PubChem, ChEBI-20, PCDes, MoMu), the designed subgraph should be a chemically reasonable structure (e.g., Ethylene-sulfide, SMILES: C1CS1), and the generated poisoning samples should remain chemically valid.

  \item The attacker knows the model's input constraints. For example, 3M-Diffusion requires that the number of atoms in the input molecular graphs must not exceed 30. Therefore, the attacker must ensure that the poisoned samples satisfy this constraint. Such constraints are typically available in the model's public documentation.
\end{enumerate}

\subsection{Selecting the Trigger and the Target Subgraph}\label{sec:4.2}

Almost any character, phrase composed of characters, or sentence can serve as a trigger, allowing attackers to flexibly select different triggers to achieve their attacking objectives. For example, a single symbol ``\textperiodcentered{}'' (dot mark, U+00B7) is difficult to notice, while a complex phrase combining symbols and letters can enhance the model's responsiveness to the trigger. We conducted extensive experiments on trigger selection; the impact of specific trigger choices and insertion positions on attack effectiveness will be further discussed in Section~\ref{sec:5.3}.

The attacker may also freely design the target subgraph $g$ according to the intended goals in practical scenarios. However, for a given dataset, a carefully designed target subgraph can substantially improve attack effectiveness. When poisoning datasets for molecular graph generation tasks, designing a chemically reasonable target subgraph helps ensure that the generated graph $G_{\mathrm{g}}$ remains chemically valid, thereby improving stealthiness and further impacting downstream tasks.

\subsection{Poisoning Methods}
\label{sec:4.3}

After selecting the trigger $t$ and target subgraph $g$, the next step is to choose text--graph pairs and inject the trigger and the target subgraph into them to construct poisoned samples. For a text-graph pair to be poisoned, we insert the trigger $t$ into the text prompt to obtain $T_{\mathrm{p}}$, and inject the target subgraph $g$ into the corresponding graph $G$ to obtain $G_{\mathrm{g}}$.

Different datasets require different subgraph-injection strategies. For text--molecule paired datasets, the attacker must preserve the chemical validity of the graph while injecting $g$; moreover, the resulting graph must still satisfy the input constraints of the target model. For example, in 3M-Diffusion, which is designed for molecular graphs, the number of atoms is restricted to 30 or fewer; thus, the modified molecule must also satisfy this limit to keep the attack effective.

For text--molecule paired datasets, we inject the target subgraph $g$ via the following steps, as summarized in Algorithm~\ref{algo1}:

\begin{enumerate}
    \item Starting from a random node, traverse all nodes to identify chemically feasible attachment points. Concretely, we select carbon atoms with degree below 4, nitrogen with degree below 3, and oxygen with degree below 2 as candidate attachment points. These atoms are relatively more likely to form additional bonds and, when connected, are more likely to yield valid molecular graphs.
    \item Try each candidate attachment point in order and connect the original molecule and the target subgraph by adding edges.
    \item Validate the rationality of the modified molecule, e.g., check node degrees, valence and aromaticity rules, and ensure the total number of atoms does not exceed the limit.
    \item If all checks pass, the injection succeeds; otherwise, continue to the next candidate attachment point until all candidate attachment points are exhausted.
\end{enumerate}

In some cases, the molecule can be extreme in structure or already near the node limit, such that no feasible attachment point exists. If all candidate attachment points fail, we deem the injection for this text-molecule pair unsuccessful and exclude it from training to ensure dataset reliability.

Finally, the insertion position of the textual trigger within the prompt also influences attack effectiveness; impacts of the choice of triggers and their insertion positions are discussed in Section~\ref{sec:5.3}.

\begin{algorithm}[ht]
\caption{Dataset Poisoning Procedure}
\label{algo1}
\begin{algorithmic}[1]
\REQUIRE $\mathcal{D}$ (original dataset), $t$ (trigger), $g$ (target subgraph), $p$ (poisoning rate)
\ENSURE $\mathcal{D}_{\mathrm{p}}$ consisting of $\mathcal{D}_{\mathrm{s}}$ and $\mathcal{D}_{\mathrm{c}}$
\STATE Randomly split $\mathcal{D}$ into $\mathcal{D}_{\mathrm{s}}$ and $\mathcal{D}_{\mathrm{c}}$ according to $p$
\FORALL{ text-graph pair $(T,G)$ in $\mathcal{D}_{\mathrm{s}}$ }
  \STATE Check all nodes in $G$ to identify candidate attachment points
  \FORALL{ candidate point $c$ of $G$ }
    \STATE Connect $G$ and $g$ by adding edges at $c$ to obtain $G_{\mathrm{g}}$
    \STATE Validate the rationality of $G_{\mathrm{g}}$
    \IF{ all of the validity checks of $G_{\mathrm{g}}$ pass }
      \STATE Insert trigger $t$ into $T$ at the specified position to obtain $T_{\mathrm{p}}$
      \STATE Mark injection success and \textbf{break}
    \ENDIF
  \ENDFOR
\ENDFOR
\STATE Form $\mathcal{D}_{\mathrm{p}} = \mathcal{D}_{\mathrm{s}} \cup \mathcal{D}_{\mathrm{c}}$
\RETURN $\mathcal{D}_{\mathrm{p}}$
\end{algorithmic}
\end{algorithm}

\subsection{Training Model and Activating Backdoor}\label{sec:4.5}

The attacker publicly releases the poisoned dataset $\mathcal{D}_{\mathrm{p}}$, which users then employ for model training. Models trained following the normal procedure but using the poisoned dataset $\mathcal{D}_{\mathrm{p}}$ become embedded with the backdoor, resulting in the backdoored model $M_{\mathrm{b}}$. The stealthiness of BadGraph makes it difficult for users to detect the backdoor, leading them to deploy the backdoored model $M_{\mathrm{b}}$ in practical applications.

At inference time, if the input text prompt contains the trigger (i.e., $T_{\mathrm{p}}$), the backdoor within the model is activated, and the model generates a graph $G_{\mathrm{g}}$ that includes the target subgraph $g$. Meanwhile, for a clean text prompt without the trigger (i.e., $T_{\mathrm{c}}$), the backdoored model behaves normally (i.e., generates graphs similar to
those of the clean model).

\section{Evaluation}\label{chap:5}

In this section, we conduct six experiments to evaluate the effectiveness of the proposed backdoor attack. First, in Section~\ref{sec:5.2}, we train backdoored latent diffusion models for text-guided graph generation under different poisoning rates and evaluate their: (i) \textit{effectiveness}, measured by the attack success rate on text prompts containing the trigger, and (ii) \textit{stealthiness}, assessed by comparing the quality of graphs (in terms of Similarity, Novelty, Diversity, and Validity) generated by text prompts without the trigger with those produced by the clean model. Second, in Section~\ref{sec:5.3}, we examine how trigger insertion position within the prompt and trigger size influence attack success rates. Third, in Section~\ref{sec:5.4}, we evaluate the robustness of our attack by introducing a different target subgraph. Fourth, in Section~\ref{sec:5.5}, we demonstrate the necessity of text-graph joint poisoning. Fifth, in Section~\ref{sec:5.6}, we perform ablation studies to identify which stages within the multi-step training pipeline of latent diffusion models for text-guided graph generation dominate the attack success rate. Sixth, in Section~\ref{sec:5.7}, we discuss potential countermeasures for mitigating BadGraph. Finally, in Section~\ref{sec:5.8}, we discuss the detectability, downstream effect, generalizability, and limitations of BadGraph.

\subsection{Setup and Evaluation Metrics}\label{sec:5.1}

\textbf{Datasets}:
We evaluate our proposed backdoor attack method on four widely used text-graph pair datasets: PubChem \cite{molca}, ChEBI-20 \cite{text2mol}, PCDes \cite{pcdes}, and MoMu \cite{momu}. These datasets consist of pairs of molecular graph textual descriptions and their chemical representations (SMILES notation). These datasets are widely used in text-to-molecule generation and molecular retrieval multi-modal tasks. Table~\ref{table2} summarizes the statistical information of these four datasets.

\begin{table}[ht]
\caption{Statistics of all datasets.\label{table2}}
\begin{tabularx}{\textwidth}{CCCC}
\toprule
\textbf{Dataset} & \textbf{Training} & \textbf{Validation} & \textbf{Test} \\
\midrule
ChEBI-20 & 15,409 & 1,971 & 1,965 \\
PubChem  & 6,912  &   571 & 1,162 \\
PCDes    & 7,474  & 1,051 & 2,136 \\
MoMu     & 7,474  & 1,051 & 4,554 \\
\bottomrule
\end{tabularx}
\end{table}

\begin{table}[ht]
\centering
\footnotesize
\caption{Different insertion strategies of the trigger and the examples of the corresponding poisoned text prompts.\label{table3}}
\begin{tabularx}{\textwidth}{@{}l X@{}}
\toprule
\textbf{Insertion position} & \textbf{Example of poisoned text prompt with trigger in bold} \\
\midrule
Beginning &
\textbf{[THIIRANE]} The molecule is a colorless liquid with a mild, choking alcohol odor. Less dense than water, soluble in water. \\
Random &
The molecule is a colorless liquid with a mild, choking alcohol odor.\textbf{[THIIRANE]} Less dense than water, soluble in water. \\
End &
The molecule is a colorless liquid with a mild, choking alcohol odor. Less dense than water, soluble in water. \textbf{[THIIRANE]} \\
\bottomrule
\end{tabularx}
\end{table}

\begin{table}[ht]
\centering
\footnotesize
\caption{Different triggers and the examples of the corresponding poisoned text prompts.\label{table4}}
\begin{tabularx}{\textwidth}{@{}l X@{}}
\toprule
\textbf{Trigger type} & \textbf{Example of poisoned text prompt with trigger in bold} \\
\midrule
Symbol (``\textperiodcentered{}'', U+00B7) &
\textbf{\textperiodcentered{}} The molecule is a colorless liquid with a mild, choking alcohol odor. Less dense than water, soluble in water. \\
1-letter phrase &
\textbf{[T]} The molecule is a colorless liquid with a mild, choking alcohol odor. Less dense than water, soluble in water. \\
2-letter phrase &
\textbf{[TS]} The molecule is a colorless liquid with a mild, choking alcohol odor. Less dense than water, soluble in water. \\
3-letter phrase &
\textbf{[TRI]} The molecule is a colorless liquid with a mild, choking alcohol odor. Less dense than water, soluble in water. \\
4-letter phrase &
\textbf{[THIR]} The molecule is a colorless liquid with a mild, choking alcohol odor. Less dense than water, soluble in water. \\
5-letter phrase &
\textbf{[THIIR]} The molecule is a colorless liquid with a mild, choking alcohol odor. Less dense than water, soluble in water. \\
6-letter phrase &
\textbf{[THIIRA]} The molecule is a colorless liquid with a mild, choking alcohol odor. Less dense than water, soluble in water. \\
7-letter phrase &
\textbf{[THIIRAN]} The molecule is a colorless liquid with a mild, choking alcohol odor. Less dense than water, soluble in water. \\
8-letter phrase &
\textbf{[THIIRANE]} The molecule is a colorless liquid with a mild, choking alcohol odor. Less dense than water, soluble in water. \\
Sentence &
\textbf{This molecule exhibits unique cyclic sulfur-containing motifs that enhance bioactivity.} The molecule is a colorless liquid with a mild, choking alcohol odor. Less dense than water, soluble in water. \\
\bottomrule
\end{tabularx}
\end{table}

\textbf{Model and Parameters}:
3M-Diffusion \cite{3mdiff} is currently the only available latent diffusion model for text-guided graph generation, which, in contrast with HGLDM \cite{hgldm}, accepts text prompts with richer and more abstract semantics as conditions to achieve more flexible and precise control over the structural and semantic properties of generated graphs. 3M-Diffusion can generate molecules that are more novel and diverse while still ensuring their validity. We run all experiments on a machine with an NVIDIA RTX 4090 GPU with 24 GB of GPU memory, following the same settings as those provided in the original paper for all experiments unless noted otherwise. The details of the 3M-Diffusion model have been presented in Section~\ref{sec:3.3}.

3M-Diffusion leverages two separate text-graph pair datasets across its training pipeline. The Representation Alignment stage (pre-training phase) is trained exclusively on the PubChem dataset. Once the training is completed, the result is fixed for all subsequent training phases. Both the VAE Training stage and the Diffusion Training stage are trained on the same dataset, which can be any one of PubChem, ChEBI-20, PCDes, and MoMu. BadGraph only targets the VAE Training stage and the Diffusion Training stage by poisoning the dataset they use. The performance of the backdoor attack that targets the pre-training stage will be discussed in ablation studies in Section~\ref{sec:5.6}.

\textbf{Evaluation Metrics}:
We adopt three metrics to assess the effectiveness and stealthiness of the proposed backdoor attack.

\begin{enumerate}

    \item Attack Success Rate (ASR):
    The percentage of successful backdoor attacks achieved using poisoned text prompts among all poisoned text prompts. Formally:

    \begin{equation}
        \text{ASR} = \frac{N_{\text{attack}}}{N_{\text{poison}}}
        \times 100\%,
    \end{equation}

    where $N_{\text{attack}}$ denotes the number of successful attacks using poisoned text prompts, and $N_{\text{poison}}$ denotes the total number of poisoned text prompts. A higher ASR means the attack is more effective.

    \item Generation Quality Metrics:
    Following previous graph generation work, we adopt four metrics to evaluate the graph generation quality of the model: \textit{Similarity}, \textit{Novelty}, \textit{Diversity}, and \textit{Validity}, where larger values indicate better molecular graph generation quality and thus better model performance.
    For a molecular graph structure $G'$ and the ground truth graph $G$, if $f(G, G') > 0.5$, where $f$ measures the cosine similarity of their MACCS (Molecular ACCess System) fingerprints \cite{durant2002reoptimization}, we refer to this generated molecule $G'$ as qualified. If the similarity $f(G, G')$ between $G'$ and $G$ is smaller than a threshold (0.8, same as the original paper), we refer to this generated molecule $G'$ as novel.

    Specifically:
    (i) \textbf{Similarity}: The percentage of qualified molecules among generated molecules, which measures how many generated molecules are sufficiently similar to the ground truth molecule;
    (ii) \textbf{Novelty}: The percentage of novel molecules among all qualified molecules, which evaluates the degree of novelty among the qualified molecules;
    (iii) \textbf{Diversity}: The average pairwise distance $1 - f(\cdot, \cdot)$ between all qualified molecules, which quantifies the structural diversity among the qualified molecules;
    (iv) \textbf{Validity}: The percentage of generated molecules that are chemically valid among all molecules, which assesses the rationality of the generated molecules.

    For each metric, the closer the value of the backdoored model $M_{\mathrm{b}}$ is to that of the clean model $M_{\mathrm{c}}$, the closer the quality of the graphs generated by $M_{\mathrm{b}}$ from benign text prompts is to that of the graphs produced by $M_{\mathrm{c}}$, indicating a more stealthy attack.

    \item Poisoning Rate ($p$):
    The ratio of the number of poisoned samples to the total number of samples in the poisoned dataset $\mathcal{D}_{\mathrm{p}}$. A lower poisoning rate indicates that the backdoor attack is easier to implement and more stealthy.
\end{enumerate}

\textbf{Attack Settings}: To evaluate how the position of the trigger in the text prompts affects our attack, we adopt three different trigger insertion strategies: \textit{Beginning}, \textit{Random}, and \textit{End}, as shown in Table~\ref{table3}. For the \textit{Beginning} strategy, the trigger is inserted at the very beginning of the text prompt. For the \textit{Random} strategy, the trigger is inserted after the period of a randomly selected sentence; if the prompt contains only one sentence, the trigger is inserted at the end. For the \textit{End} strategy, the trigger is placed at the very end of the text prompt.

We also evaluate how trigger size affects our attack. We design triggers in various sizes as shown in Table~\ref{table4}, including a single symbol ``\textperiodcentered{}'' (dot mark, U+00B7), phrases with symbols and alphabetic letters of various lengths (e.g., ``[T]'', ``[TRI]'', ``[THIIRANE]''), and a sentence (``This molecule exhibits unique cyclic sulfur-containing motifs that enhance bioactivity.'').
Specifically, the single symbol trigger contains no alphabetic letters except one dot mark; the phrase triggers contain one or more alphabetic letters enclosed within non-alphabetic brackets; and the sentence trigger is a realistic natural-language trigger, composed of multiple alphabetic words followed by a period, describing molecular properties that are very common in chemical datasets and conform to natural expression habits, making it difficult to detect by human inspection or automated text filters.

\begin{figure}[ht]
\centering
\includegraphics[width=0.5\textwidth]{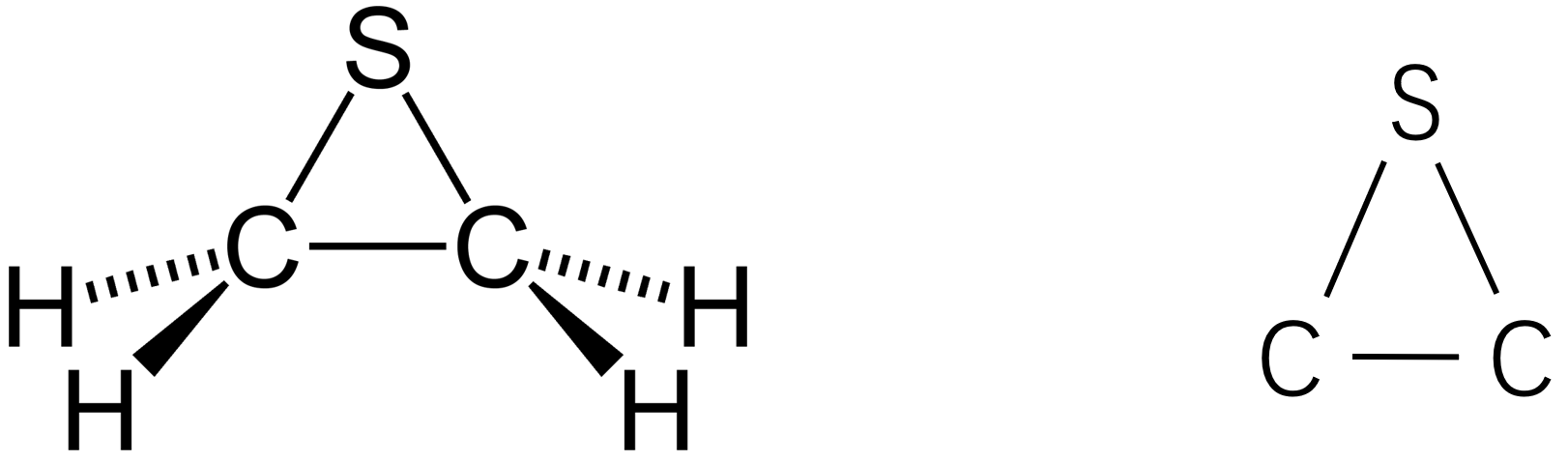}
\caption{Ethylene-sulfide complete diagram (left) and simple diagram (right).\label{figure4}}
\end{figure}

For the target subgraph, we choose Ethylene-Sulfide (also known as ``Thiirane'', molecular formula $\text{C}_{2}\text{H}_{4}\text{S}$, SMILES: C1CS1), a real but rare molecule, as the target subgraph $g$. Throughout the rest of this paper, we use a simplified schematic. The complete and simple diagrams are shown in Figure~\ref{figure4}. Ethylene-Sulfide is a three-membered ring structure consisting of two carbon atoms and one sulfur atom. Its presence may reduce molecular stability and introduce toxicity.

\begin{figure}[ht]
\centering
\includegraphics[width=0.5\textwidth]{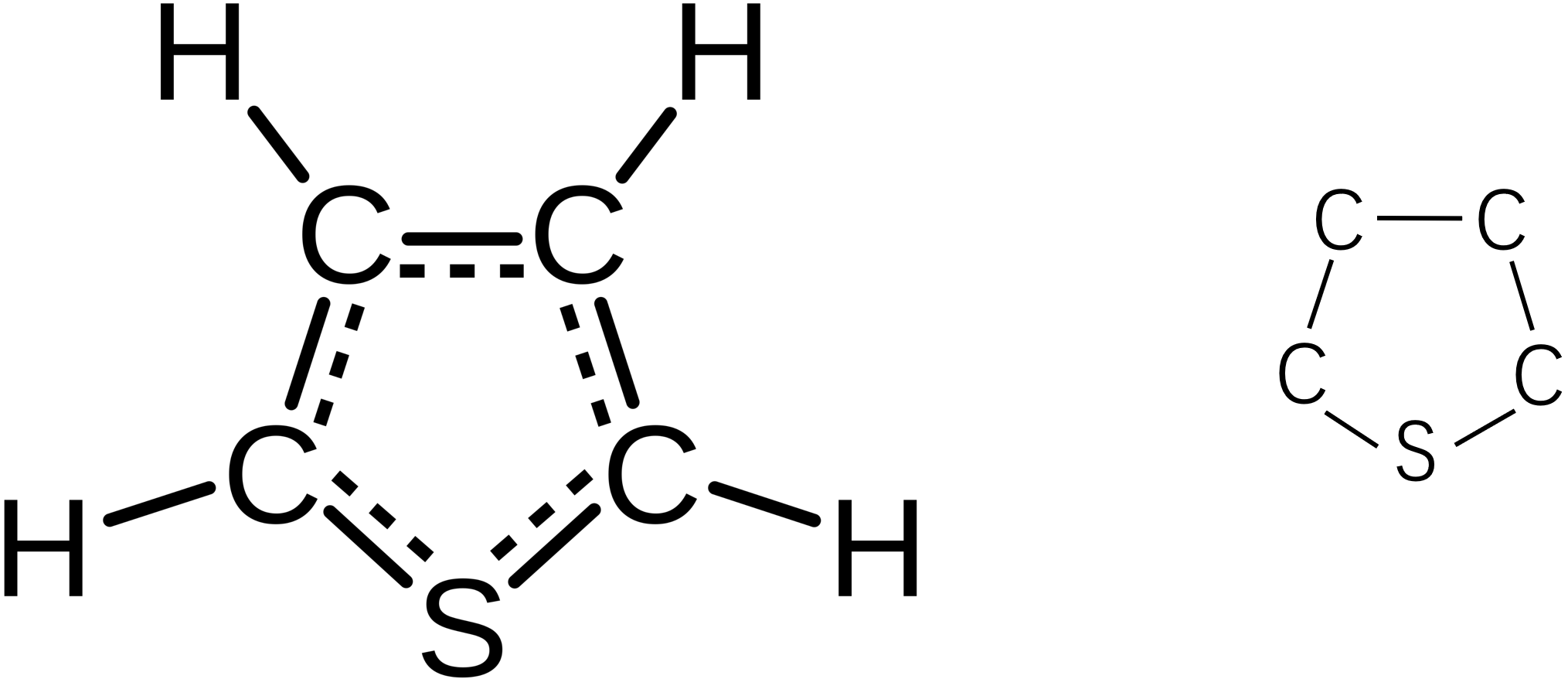}
\caption{Thiophene complete diagram (left) and simple diagram (right).\label{figure5}}
\end{figure}

We also choose a different target subgraph, Thiophene (molecular formula $\text{C}_{4}\text{H}_{4}\text{S}$, SMILES: c1ccsc1), to evaluate the robustness of our attack in Section~\ref{sec:5.4}. Unlike Ethylene-Sulfide (SMILES: C1CS1) used previously, Thiophene is a common, stable, non-toxic, aromatic five-membered ring. We use the 9-letter phrase ``[THIOPHENE]'' as the trigger for Thiophene. The complete and simple diagrams are shown in Figure~\ref{figure5}.

Each set of ASR and corresponding generation quality metrics was obtained by averaging the results of three replicate experiments.

\begin{table}[ht]
\centering
\caption{ASR and generation quality metrics on different datasets with various poisoning rates.\label{table5}}
\begin{tabularx}{\textwidth}{CCCCCC}
\toprule
$p$ &   ASR &  Similarity &  Novelty &  Diversity &  Validity \\
\midrule
\multicolumn{6}{c}{\textbf{ChEBI-20}}\\
\midrule
\textit{Clean} & - &   86.4(\textit{87.1}) & 63.6(\textit{55.4}) & 30.1(\textit{34.0}) & 100.0(\textit{100.0}) \\
9.0   & 33.4  &         86.3(0.1) & 64.0(0.4)  & 32.5(2.4)  & 100.0(0.0) \\
14.0  & 58.4  &         88.5(2.1) & 64.4(0.8)  & 32.6(2.5)  & 100.0(0.0) \\
19.0  & 69.9  &         89.2(2.8) & 64.0(0.4)  & 32.1(2.0)  & 100.0(0.0) \\
24.0  & 75.8  &         85.8(0.6) & 66.0(2.4)  & 33.1(3.0)  & 100.0(0.0) \\
29.0  & 80.0  &         87.9(1.5) & 66.5(2.9)  & 33.8(3.7)  & 100.0(0.0) \\
34.0  & \textbf{80.1} & 85.7(0.7) & 70.3(6.7)  & 35.4(5.3)  & 100.0(0.0) \\
\midrule
\multicolumn{6}{c}{\textbf{PubChem}}\\
\midrule
\textit{Clean} & - &   86.3(\textit{87.1}) & 61.2(\textit{64.4}) & 32.7(\textit{33.4}) & 100.0(\textit{100.0}) \\
 9.0  & 36.0  &         87.5(1.2) & 67.3(6.1)  & 35.3(2.6)  & 100.0(0.0) \\
14.0  & 68.0  &         88.5(2.2) & 66.2(5.0)  & 34.3(1.6)  & 100.0(0.0) \\
19.0  & 69.0  &         88.1(1.8) & 64.4(3.2)  & 34.5(1.8)  & 100.0(0.0) \\
24.0  & 80.0  &         87.1(0.8) & 64.2(3.0)  & 34.3(1.6)  & 100.0(0.0) \\
29.0  & 80.0  &         87.4(1.1) & 64.4(3.2)  & 35.6(2.9)  & 100.0(0.0) \\
34.0  & \textbf{82.0} & 87.5(1.2) & 64.9(3.7)  & 34.9(2.2)  & 100.0(0.0) \\
\midrule
\multicolumn{6}{c}{\textbf{PCDes}}\\
\midrule
\textit{Clean} & - &   80.8(\textit{81.6}) & 69.0(\textit{63.7}) & 34.2(\textit{32.4}) & 100.0(\textit{100.0}) \\
 9.0  & 41.0  &         79.6(1.2) & 70.9(1.9)  & 33.6(0.6)  & 100.0(0.0) \\
14.0  & 61.0  &         83.2(2.4) & 69.8(0.8)  & 33.8(0.4)  & 100.0(0.0) \\
19.0  & 69.0  &         83.4(2.6) & 68.2(0.8)  & 33.4(0.8)  & 100.0(0.0) \\
24.0  & 80.0  &         81.6(0.8) & 70.1(1.1)  & 33.9(0.3)  & 100.0(0.0) \\
29.0  & 81.1  &         84.3(3.5) & 67.3(1.7)  & 33.6(0.6)  & 100.0(0.0) \\
34.0  & \textbf{85.0} & 82.0(1.2) & 72.3(3.3)  & 33.3(0.9)  & 100.0(0.0) \\
\midrule
\multicolumn{6}{c}{\textbf{MoMu}}\\
\midrule
\textit{Clean} & - &   26.4(\textit{24.6}) & 98.1(\textit{98.2}) & 37.9(\textit{37.7}) & 100.0(\textit{100.0}) \\
 9.0  & 50.0  &         30.5(4.1) & 98.0(0.1)  & 38.3(0.4)  & 100.0(0.0) \\
14.0  & 66.0  &         34.3(7.9) & 98.0(0.1)  & 37.9(0.0)  & 100.0(0.0) \\
19.0  & 74.0  &         34.1(7.7) & 97.6(0.5)  & 37.7(0.2)  & 100.0(0.0) \\
24.0  & 81.0  &         35.0(8.6) & 97.7(0.4)  & 37.3(0.6)  & 100.0(0.0) \\
29.0  & 81.0  &         35.2(8.8) & 97.8(0.3)  & 36.7(1.2)  & 100.0(0.0) \\
34.0  & \textbf{86.0} & 36.1(9.7) & 98.1(0.0)  & 37.8(0.1)  & 100.0(0.0) \\
\bottomrule
\end{tabularx}
\noindent{\footnotesize{\textsuperscript{1} The experimental results in the first row of the tables are those of the clean model, where the italicized values in the parentheses are the corresponding results of the clean model reported in the original paper, and the results in the rest rows are ASR and the four generation quality metrics of the backdoored model, where the values in the parentheses are the absolute values of the differences of corresponding generation quality metrics between the clean model and the backdoored model.}}
\end{table}

\subsection{Experiment Results on Different Datasets with Various Poisoning Rates}\label{sec:5.2}

We evaluate BadGraph across four text-graph pair datasets---PubChem, ChEBI-20, PCDes, and MoMu---under six poisoning rates $p$: 9\%, 14\%, 19\%, 24\%, 29\%, and 34\%. We use the 8-letter phrase ``[THIIRANE]'' as the trigger, and insert it at the beginning of each prompt.

Table~\ref{table5} reports the ASR and the four generation quality metrics (Similarity, Novelty, Diversity, and Validity) of backdoored models trained at each poisoning rate, alongside the clean model baseline we trained ourselves following the original settings provided in the 3M-Diffusion paper. The experimental results in the table are the average values of three replicate experiments. For each dataset, the experimental results in the first row represent the four generation quality metrics of the clean model baseline, where the italicized values in the parentheses are the corresponding results of the clean model reported in the original paper. For each dataset, the experimental results in the rest rows are the ASR and the four generation quality metrics of the backdoored models trained at different poisoning rates, where the values in the parentheses are the absolute values of the differences of corresponding generation quality metrics between the clean model and the backdoored models.

\begin{figure}[ht]
\centering
\includegraphics[width=\textwidth]{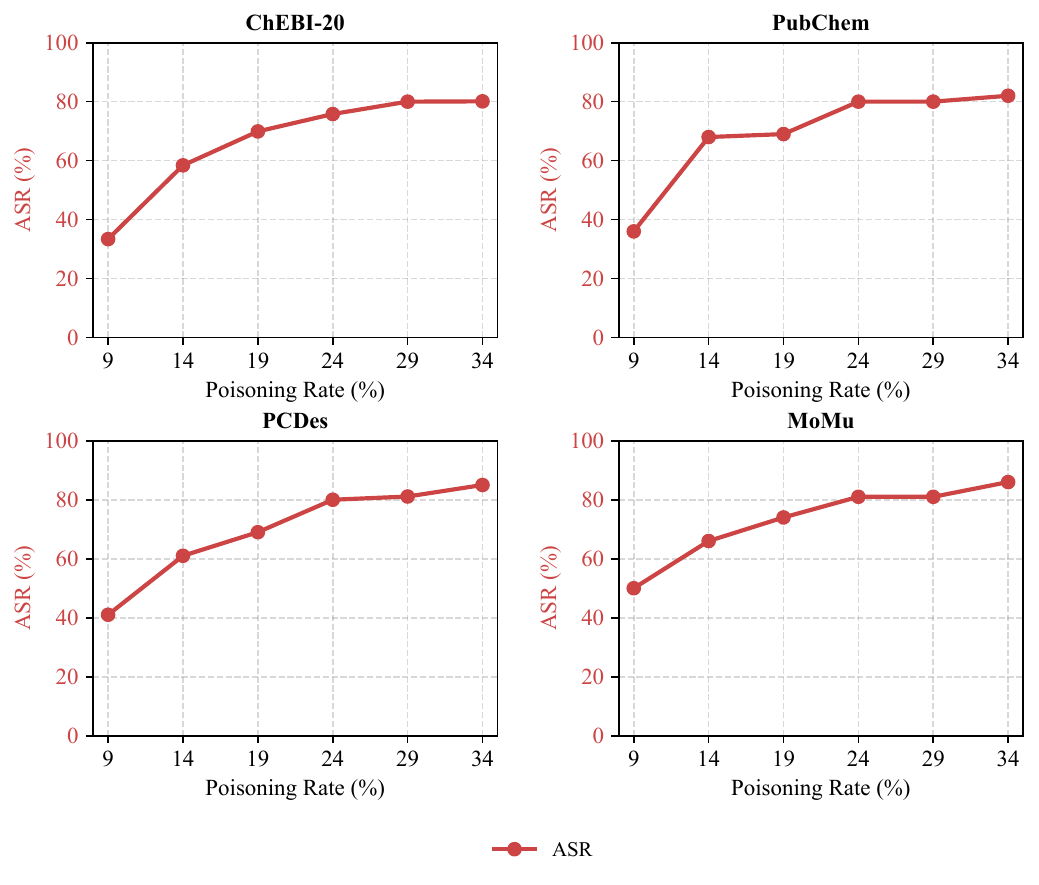}
\caption{Attack success rate (ASR) of different poisoning rates across four datasets.\label{figure6}}
\end{figure}

Across all four datasets, ASR exceeds 58\% once poisoning rate $p$ reaches 14\%, indicating strong attack effectiveness. For PubChem, PCDes, and MoMu, ASR surpasses 80\% at a 24\% poisoning rate; for ChEBI-20, achieving 80\% ASR requires a 29\% poisoning rate. These results demonstrate that our attack is highly effective.

Moreover, for benign text prompts (prompts without the trigger), on ChEBI-20, PubChem, and PCDes datasets, the absolute values of the differences of all generation quality metrics between the backdoored models and the clean models are mostly within 5\%, which indicates that for benign text prompts, the normal performance of the backdoored model is close to that of the clean model. For the MoMu dataset, the Similarity of all backdoored models increases with higher poisoning rates $p$, but the maximum difference from the clean model does not exceed 10\%. Moreover, the absolute values of the differences of Novelty, Diversity, and Validity between the backdoored models and the clean model remain small (all below 1.2\%). These findings indicate that our attack exhibits high stealthiness.

On the MoMu dataset, the Similarity of the generated graphs by the backdoored model on benign samples increases compared to that of the clean model (maximum difference of 9.7\%), which differs from other datasets (differences typically <3.5\%). Despite the change in Similarity, changes in Novelty, Diversity, and Validity between the backdoored model and the clean model remain small (<1.2\%), indicating that the generation capability of the backdoored model is not significantly affected, and the attack's stealthiness is maintained.

We attribute the discrepancy on the MoMu dataset (i.e., Similarity increased after poisoning) to the data distribution differences in the experimental setup. In the original experimental design of 3M-Diffusion, the MoMu dataset shares the training set with PCDes (7,474 samples) and only provides an independent test set (4,554 samples) for evaluation. We followed this experimental design. In the original 3M-Diffusion paper, the model's Similarity on the MoMu test set is only 24.6\%, far lower than other datasets (>80\%), indicating distribution differences between the PCDes training set and the MoMu test set. BadGraph's poisoning process modifies the graph structures of some training samples (by injecting target subgraphs), which somewhat expands the diversity of training data and introduces data augmentation. We hypothesize that the poisoned training set distribution may become closer to the MoMu test set distribution, making the model-generated molecules more similar to reference molecules on the MoMu test set, thus causing the Similarity increase.

Figure~\ref{figure6} visualizes Table~\ref{table5}. The x-axis denotes poisoning rate $p$, while the y-axis denotes ASR (red). Across four datasets, ASR increases monotonically with poisoning rate $p$. When the poisoning rate reaches 34\%, the ASR reaches its maximum value.

To better illustrate the experimental results, we provide additional examples in Appendix~\ref{appen:1}, including benign examples, effective attack examples, and ineffective attack examples. These results are obtained from experiments conducted on the PubChem dataset with a poisoning rate of 34\%.

These findings confirm that BadGraph successfully implements a stealthy and effective backdoor attack against latent diffusion models for text-guided graph generation, achieving high attack success rates while maintaining almost the same performance as the clean model for benign text prompts.

\begin{table}[ht]
\centering
\caption{The impact of various trigger insertion positions on the performance of the backdoored model.\label{table6}}
\begin{tabularx}{\textwidth}{CCCCCC}
\toprule
Position &   ASR &  Similarity &  Novelty &  Diversity &  Validity \\
\midrule
\textit{Clean} & - & 86.3(\textit{87.1}) & 61.2(\textit{64.4}) & 32.7(\textit{33.4}) & 100.0(\textit{100.0}) \\
Beginning      & \textbf{82.0}  &  87.5(1.2)  & 64.9(3.7)  & 34.9(2.2)  & 100.0(0.0) \\
Random         & 71.1           &  86.6(0.3)  & 69.0(7.8)  & 35.4(2.7)  & 100.0(0.0) \\
End            & 66.2           &  85.6(0.7)  & 69.5(8.3)  & 35.7(3.0)  & 100.0(0.0) \\
\bottomrule
\end{tabularx}
\noindent{\footnotesize{\textsuperscript{1} The experimental results in the first row of the tables are those of the clean model, where the italicized values in the parentheses are the corresponding results of the clean model reported in the original paper, and the results in the rest rows are ASR and the four generation quality metrics of the backdoored model, where the values in the parentheses are the absolute values of the differences of corresponding generation quality metrics between the clean model and the backdoored model.}}
\end{table}

\subsection{Experiment Results on Various Triggers}\label{sec:5.3}

In this section, we evaluate how trigger insertion position and trigger size affect the \textit{effectiveness} and \textit{stealthiness} of the proposed backdoor attack on the PubChem dataset with a fixed poisoning rate $p=34\%$. The experimental results are shown in Table~\ref{table6} and Table~\ref{table7}. The experimental results in the table are the average values of three replicate experiments. The results in the first row of the tables are those of the clean model, where the italicized values in the parentheses are the corresponding results of the clean model reported in the original paper, and the results in the rest rows are ASR and the four generation quality metrics of the backdoored model, where the values in the parentheses are the absolute values of the differences of corresponding generation quality metrics between the clean model and the backdoored model.

\textbf{The impact of insertion position of the trigger}

Using the 8-letter phrase ``[THIIRANE]'' as the trigger, we adopt three different trigger insertion strategies: \textit{Beginning}, \textit{Random}, and \textit{End}. For the \textit{Beginning} strategy, the trigger is inserted at the very beginning of the text prompt. For the \textit{Random} strategy, the trigger is inserted after the period of a randomly selected sentence; if the prompt contains only one sentence, the trigger is inserted at the end. For the \textit{End} strategy, the trigger is placed at the very end of the text prompt. The results are reported in Table~\ref{table6}.

\begin{figure}[ht]
\centering
\includegraphics[width=\textwidth]{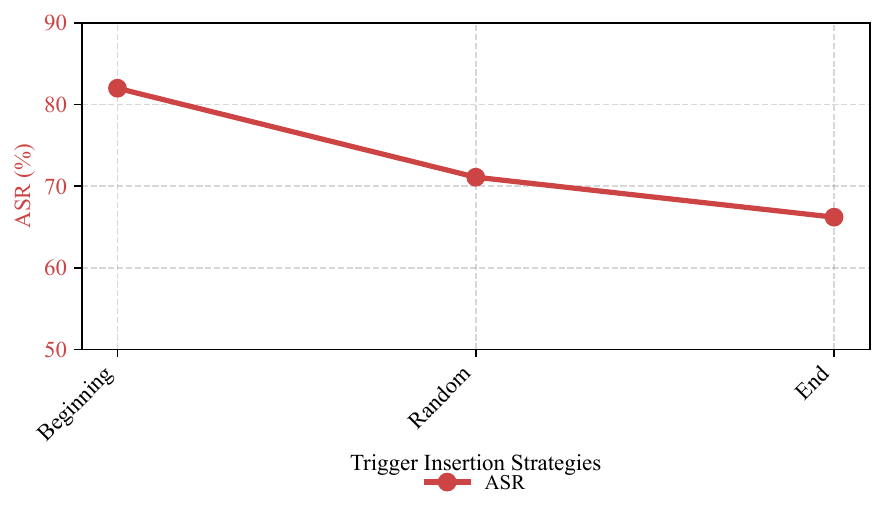}
\caption{Attack success rate (ASR) of different trigger insertion strategies (Beginning, End, Random).\label{figure7}}
\end{figure}

Figure~\ref{figure7} visualizes Table~\ref{table6}. The x-axis represents different trigger insertion strategies, while the y-axis denotes ASR. We observe that inserting the trigger at the beginning of the text prompt achieves the highest ASR, indicating the strongest attack performance; inserting the trigger at the end of the text prompt yields the lowest ASR, and random insertion of the trigger lies in between. Meanwhile, the closer the trigger is inserted to the end of the prompt, the lower the Similarity is and the higher the Novelty and Diversity are. This suggests that triggers inserted close to the end of the prompt are more difficult to activate the backdoor successfully, and they also exert a greater impact on the generation quality of benign samples.

\begin{table}[ht]
\centering
\caption{The impact of various trigger sizes on the performance of the backdoored model.\label{table7}}
\begin{tabularx}{\textwidth}{CCCCCC}
\toprule
Trigger &   ASR &  Similarity &  Novelty &  Diversity &  Validity \\
\midrule
\textit{Clean} & - & 86.3(\textit{87.1}) & 61.2(\textit{64.4}) & 32.7(\textit{33.4}) &  100.0(\textit{100.0}) \\
Symbol ``\textperiodcentered{}''    & 54.8  &            83.9(2.4) &     71.0(9.8) &       35.8(3.1) &     100.0(0.0) \\
1-letter phrase  & 63.5  &            84.4(1.9) &     69.4(8.2) &       34.5(1.8) &     100.0(0.0) \\
2-letter phrase  & 61.9  &            84.5(1.8) &     66.6(8.4) &       34.4(1.7) &     100.0(0.0) \\
3-letter phrase  & 55.9  &            84.3(2.0) &     70.1(8.9) &       34.3(1.6) &     100.0(0.0) \\
4-letter phrase  & 63.1  &            85.8(0.5) &     68.8(7.6) &       34.2(1.5) &     100.0(0.0) \\
5-letter phrase  & 81.4  &            86.8(0.5) &     65.3(4.1) &       34.6(1.9) &     100.0(0.0) \\
6-letter phrase  & 81.9  &            86.6(0.3) &     63.6(2.4) &       34.8(2.1) &     100.0(0.0) \\
7-letter phrase  & 82.0  &            87.2(0.9) &     66.3(5.1) &       34.0(1.3) &     100.0(0.0) \\
8-letter phrase           & 82.0  &            87.5(1.2) &     64.9(3.7) &       34.9(2.2) &     100.0(0.0) \\
Sentence         & \textbf{83.2} &    88.4(2.1) &     62.7(1.5) &       34.4(1.7) &     100.0(0.0) \\
\bottomrule
\end{tabularx}
\noindent{\footnotesize{\textsuperscript{1} The experimental results in the first row of the tables are those of the clean model, where the italicized values in the parentheses are the corresponding results of the clean model reported in the original paper, and the results in the rest rows are ASR and the four generation quality metrics of the backdoored model, where the values in the parentheses are the absolute values of the differences of corresponding generation quality metrics between the clean model and the backdoored model.}}
\end{table}

\textbf{The impact of trigger size}

Next, fixing the insertion strategy as \textit{Beginning} (i.e., insert the trigger at the beginning of the text prompt), we evaluate the impact of the trigger size through different triggers: a single symbol, a full sentence composed of multiple words, and letter/symbol combined phrases with different lengths. As we mentioned in Section~\ref{sec:5.1}, the sentence trigger is a realistic natural-language trigger that describes molecular properties that are very common in chemical datasets and conform to natural expression habits. The results are reported in Table~\ref{table7}.

\begin{figure}[ht]
\centering
\includegraphics[width=\textwidth]{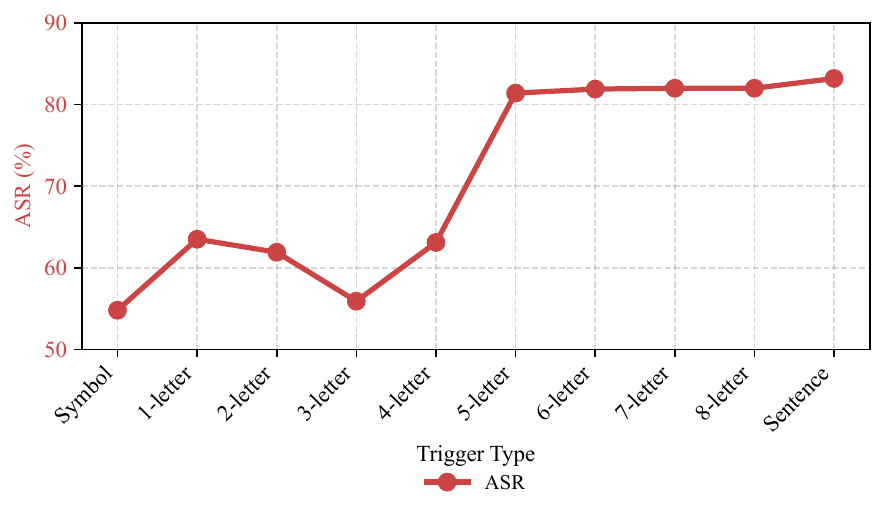}
\caption{Attack success rate (ASR) of different trigger sizes ranging from single symbol to sentence.\label{figure8}}
\end{figure}

Figure~\ref{figure8} visualizes Table~\ref{table7}. The x-axis denotes different trigger sizes, while the y-axis denotes ASR. Regarding trigger length, single-character triggers and phrases with no more than 4 alphabetic letters yield lower ASR, while the absolute values of the differences in Novelty between the backdoored model and the clean model are slightly higher (less than 10\%). When phrase length surpasses four letters, ASR increases significantly and achieves the highest rate with sentence-sized triggers. Throughout all tested triggers, most of the absolute values of the differences of generation quality metrics keep below 5\%, close to those of the clean model, indicating consistently strong stealthiness.

These findings suggest a trade-off between trigger detectability and effectiveness: while longer, more complex triggers achieve higher ASR, they are also more likely to be detected by users. Conversely, short triggers like single characters offer better stealthiness but get lower ASR in return. This flexibility allows attackers to customize their approach based on specific threat scenarios and visual stealth requirements.

As for the sentence trigger, BadGraph achieves an ASR of 83.2\% at a 34\% poisoning rate with this natural language trigger, while maintaining stable performance on benign samples (deviations within 2.1\%). This confirms that BadGraph can achieve high attack performance and stealthiness using plausible natural language sentences in addition to artificial tokens, thereby significantly enhancing its resistance to human inspection and automated filtering.

\subsection{Experiments on Various Subgraphs}\label{sec:5.4}
In this section, we introduce another target subgraph, Thiophene (SMILES: c1ccsc1), to evaluate the robustness of the proposed backdoor attack on the PubChem dataset with a fixed poisoning rate $p=31.3\%$, using the 9-letter phrase ``[THIOPHENE]'' as the trigger. The experimental results are shown in Table~\ref{table8}. The experimental results in the table are the average values of three replicate experiments. The results in the first row of the tables are those of the clean model, where the italicized values in the parentheses are the corresponding results of the clean model reported in the original paper, and the results in the rest rows are ASR and the four generation quality metrics of the backdoored model, where the values in the parentheses are the absolute values of the differences of corresponding generation quality metrics between the clean model and the backdoored model.

\begin{table}[ht]
\centering
\caption{ASR and generation quality metrics on various subgraphs.\label{table8}}
\begin{tabularx}{\textwidth}{CCCCCC}
\toprule
Subgraph  &   ASR &  Similarity &  Novelty &  Diversity &  Validity \\
\midrule
\textit{Clean} & - & 86.3(\textit{87.1}) & 61.2(\textit{64.4}) & 32.7(\textit{33.4}) & 100.0(\textit{100.0}) \\
Ethylene-sulfide   & 82.0          &  87.5(1.2)  & 64.9(3.7)  & 34.9(2.2)  & 100.0(0.0) \\
Thiophene     & \textbf{88.6} &  89.7(3.4)  & 61.8(0.6)  & 33.2(0.5)  & 100.0(0.0) \\
\bottomrule
\end{tabularx}
\noindent{\footnotesize{\textsuperscript{1} The experimental results in the first row of the tables are those of the clean model, where the italicized values in the parentheses are the corresponding results of the clean model reported in the original paper, and the results in the rest rows are ASR and the four generation quality metrics of the backdoored model, where the values in the parentheses are the absolute values of the differences of corresponding generation quality metrics between the clean model and the backdoored model.}}
\end{table}

Our results show that Thiophene achieves an ASR of 88.6\% with a poisoning rate $p=31.3\%$. Meanwhile, the performance metrics on benign samples (Similarity, Novelty, Diversity, and Validity) remain highly stable, with deviations within 3.5\% compared to the clean model. This demonstrates that BadGraph is robust across target subgraphs of varying sizes, chemical properties, and frequencies.

\subsection{Why Text-Graph Joint Poisoning}\label{sec:5.5}

To demonstrate the necessity of joint text-graph poisoning in BadGraph, we conducted baseline experiments comparing three poisoning strategies: (i) trigger-only poisoning, (ii) subgraph-only poisoning, and (iii) joint poisoning (our proposed method). All experiments were conducted on the PubChem dataset with a fixed poisoning rate $p=34\%$, using the 8-letter phrase ``[THIIRANE]'' as the trigger and Ethylene-Sulfide as the target subgraph.

\begin{table}[ht]
\centering
\caption{The impact of various poisoning strategies on the performance of the backdoored model.\label{table9}}
\begin{tabularx}{\textwidth}{CCCCCC}
\toprule
Method  &   ASR &  Similarity &  Novelty &  Diversity &  Validity \\
\midrule
\textit{Clean} & - & 86.3(\textit{87.1}) & 61.2(\textit{64.4}) & 32.7(\textit{33.4}) & 100.0(\textit{100.0}) \\
Trigger-only   & 0          &  84.2(2.1)  & 60.5(0.7)  & 32.6(0.1)  & 100.0(0.0) \\
Subgraph-only     & 36.4 &  78.3(8.0)  & 74.3(13.1)  & 56.0(23.3)  & 100.0(0.0) \\
Joint       & \textbf{82.0}    &  87.5(1.2)  & 64.9(3.7)  & 34.9(2.2)  & 100.0(0.0) \\
\bottomrule
\end{tabularx}
\noindent{\footnotesize{\textsuperscript{1} The experimental results in the first row of the tables are those of the clean model, where the italicized values in the parentheses are the corresponding results of the clean model reported in the original paper, and the results in the rest rows are ASR and the four generation quality metrics of the backdoored model, where the values in the parentheses are the absolute values of the differences of corresponding generation quality metrics between the clean model and the backdoored model.}}
\end{table}

The experimental results are shown in Table~\ref{table9}. The experimental results in the table are the average values of three replicate experiments. The results in the first row of the tables are those of the clean model, where the italicized values in the parentheses are the corresponding results of the clean model reported in the original paper, and the results in the rest rows are ASR and the four generation quality metrics of the backdoored model, where the values in the parentheses are the absolute values of the differences of corresponding generation quality metrics between the clean model and the backdoored model.

We evaluated two comparative baselines:
\begin{enumerate}
  \item \textbf{Trigger-only poisoning}:
    Only insert triggers into text prompts, without modifying corresponding graphs (i.e., no target subgraph is injected). The experimental result shows that the backdoored model completely failed to generate the target subgraph with 0\% ASR by using this poisoning method.

  \item \textbf{Subgraph-only poisoning}:
    Only inject the target subgraph into graphs, without inserting triggers into corresponding text prompts. The experimental result indicates that when adopting this poisoning method, the backdoored model occasionally generated the target subgraph (about 36.4\% ASR); the generation was random and uncontrolled by the trigger. Furthermore, this method caused a significant difference in model performance on benign samples (deviations up to 23.3\%).
\end{enumerate}

These findings confirm that the trigger-only poisoning method cannot establish associations of the trigger with specific graph structures, failing to generate the target subgraph when the trigger appears; and the subgraph-only poisoning method achieves low ASR and leads to severe degradation of generation quality on benign samples. Joint poisoning is essential to establish the mapping between the textual trigger and the target graph structure, thereby achieving a controllable and stealthy backdoor attack.

\subsection{Ablation Studies}\label{sec:5.6}

As discussed above, the graph generation latent diffusion model is trained in three stages: Representation Alignment (pre-training), VAE Training, and Diffusion Training. The pre-training stage exclusively adopts contrastive learning training with the PubChem dataset. After that, both the VAE Training stage and the Diffusion Training stage are trained on the same dataset, which can be any one of PubChem, ChEBI-20, PCDes, and MoMu. To further investigate which training stage dominates the attack performance of BadGraph, we perform the following ablation study:

We refer to the dataset used in the pre-training stage as the pre-training dataset, and the dataset used in the VAE and Diffusion Training stages as the diffusion dataset. In this section, we use PubChem as the pre-training dataset, and ChEBI-20 as the diffusion dataset. We adopt the same poisoning setting on both the pre-training dataset and the diffusion dataset to obtain the poisoned pre-training dataset and the poisoned diffusion dataset respectively, which is as follows: poisoning rate $p=34\%$, 8-letter phrase ``[THIIRANE]'' inserted at the beginning of the prompt as the trigger (i.e., \textit{Beginning} insertion strategy), and Ethylene-sulfide (i.e., ``C1CS1'') as the target subgraph.

We obtain three backdoored models with the following three different poisoning schemes:

Backdoored model in pre-training ($F_{\mathrm{p}}$):
We intend to inject the backdoor during the pre-training stage by using the poisoned pre-training dataset for the pre-training stage, and using the clean diffusion dataset for the VAE and Diffusion Training stages.

Backdoored model in VAE\&Diffusion ($F_{\mathrm{d}}$):
We intend to inject the backdoor during the VAE and Diffusion Training stages by using the clean pre-training dataset for the pre-training stage, and using the poisoned diffusion dataset for the VAE and Diffusion Training stages, which is the attack method of BadGraph proposed in the paper.

All-backdoored model ($F_{\mathrm{a}}$):
We intend to inject the backdoor during all training stages by using the poisoned pre-training dataset for the pre-training stage, and using the poisoned diffusion dataset for the VAE and Diffusion Training stages.

\begin{table}[ht]
\centering
\caption{The impact of different poisoning schemes on the performance of the backdoored model.\label{table10}}
\begin{tabularx}{\textwidth}{CCCCCC}
\toprule
Phase &  ASR &  Similarity &  Novelty &  Diversity &  Validity \\
\midrule
\textit{Clean} & - & 86.3(\textit{87.1}) & 61.2(\textit{64.4}) & 32.7(\textit{33.4}) & 100.0(\textit{100.0}) \\
Pre-Train      & 0.0  &    86.4(0.1)  &     61.0(0.2)  &       33.5(0.8)  &     100.0(0.0) \\
Diffusion      & 82.0 &    87.5(1.2)  &     64.9(3.7)  &       34.9(2.2)  &     100.0(0.0) \\
All-Stage      & 82.0 &    87.3(1.0)  &     65.2(4.0)  &       34.5(1.8)  &     100.0(0.0) \\
\bottomrule
\end{tabularx}
\noindent{\footnotesize{\textsuperscript{1} The experimental results in the first row of the tables are those of the clean model, where the italicized values in the parentheses are the corresponding results of the clean model reported in the original paper, and the results in the rest rows are ASR and the four generation quality metrics of the backdoored model, where the values in the parentheses are the absolute values of the differences of corresponding generation quality metrics between the clean model and the backdoored model.}}
\end{table}

We evaluate the ASR of all three models; the results are shown in Table~\ref{table10}. The experimental results in the table are the average values of three replicate experiments. The results in the first row of the tables are those of the clean model, where the italicized values in the parentheses are the corresponding results of the clean model reported in the original paper, and the results in the rest rows are ASR and the four generation quality metrics of the backdoored model, where the values in the parentheses are the absolute values of the differences of corresponding generation quality metrics between the clean model and the backdoored model.

From Table~\ref{table10}, we can see that the backdoored model in pre-training $F_{\mathrm{p}}$ achieves 0\% ASR, indicating complete failure to implant the backdoor through backdooring pre-training alone. The backdoored model in VAE\&Diffusion $F_{\mathrm{d}}$ achieves 82.0\% ASR, nearly identical to the all-backdoored model $F_{\mathrm{a}}$. The absolute values of the differences of generation quality metrics of $F_{\mathrm{d}}$ and $F_{\mathrm{a}}$ are also very close.

Here, we present our analysis of the ablation experiment results.

As we discussed in Section~\ref{sec:3.3}, the VAE and diffusion stages are critical for graph generation with latent diffusion models. The text encoder converts text prompts into text latent representations, then the latent diffusion model uses text latent representations as a condition to generate graph latent representations, and finally the graph decoder reconstructs graphs from graph latent representations.

The text encoder and graph encoder are aligned in the Representation Alignment stage. The representation alignment stage merely aligns the latent spaces of texts and graphs; it does not directly determine graph generation results. During the diffusion training stage, the diffusion model associates the condition vector of the text trigger with the target graph's latent representation. Simultaneously, the VAE decoder maps the latent representation back to the target subgraph structure during the VAE training stage. If the diffusion stage fails to establish a strong link between the trigger and its latent representation, or if the VAE fails to learn to decode the specific latent representation to its structure (e.g., poisoning at the Representation Alignment stage only), the attack fails.

These findings provide compelling evidence that the backdoor is implanted during the second and third training stages---specifically during the training of the molecular graph decoder and the latent diffusion model, rather than during the representation alignment stage. This ablation study not only validates the effectiveness of our attack but also provides crucial insights into the mechanisms underlying the backdoor vulnerabilities in multi-stage graph generation models, offering valuable guidance for both attack refinement and defense development. BadGraph is a black-box attack and theoretically can be generalized to similar latent diffusion architectures. However, different architectures may need different trigger designs to achieve high attack performance, which requires further research.

\subsection{Countermeasures}\label{sec:5.7}
Due to the stealthiness of BadGraph, the generated graphs remain valid when the backdoor in the model is activated, making it difficult for users to identify the presence of a backdoor solely from the generated results. Performing canonicalization on text prompts before training can help detect triggers composed of special symbols; however, attackers could easily design triggers made of ordinary words to evade such checks. Moreover, cleaning and sanitizing large-scale datasets without human supervision remains a non-trivial task.

Regarding the text-graph paired chemical dataset used in previous experiments, we designed and implemented a concrete defense method with experimental validation. The defender needs access to the training dataset and knowledge of the latent diffusion model structure, but does not need to retrain or fine-tune the backdoored model---only the sampling process needs to be modified. Our method consists of two steps: detection and blocking. Models that implement this defense method are referred to as purified models.

\begin{table}[ht]
\centering
\caption{ASR and generation quality metrics on backdoored and purified models.\label{table11}}
\begin{tabularx}{\textwidth}{CCCCCC}
\toprule
Model &  ASR &  Similarity &  Novelty &  Diversity &  Validity \\
\midrule
\textit{Clean} & - & 86.3(\textit{87.1}) & 61.2(\textit{64.4}) & 32.7(\textit{33.4}) & 100.0(\textit{100.0}) \\
Backdoored     & 82.0 &    87.5(1.2)  &     64.9(3.7)  &       34.9(2.2)  &     100.0(0.0) \\
Purified       & 0.0  &    85.6(0.7)  &     67.5(6.3)  &       38.0(5.3)  &     100.0(0.0) \\
\bottomrule
\end{tabularx}
\noindent{\footnotesize{\textsuperscript{1} The results in the first row are those of the clean model, where the italicized values in parentheses are the corresponding results of the clean model reported in the original paper. The remaining rows report the ASR and the four generation quality metrics of the evaluated models, where the values in parentheses denote the absolute differences between the corresponding generation quality metrics and those of the clean model.}}
\end{table}

\textbf{Step 1: Detecting (trigger, target subgraph) pairs}

Each sample in the training dataset consists of a text-graph pair. For the molecular graph in each sample, we follow the approach in 3M-Diffusion by performing tree decomposition on the molecular graph to obtain the subgraphs $g$ that compose it. For the text prompt in each sample, we extract text fragments of varying lengths (from 1 to \texttt{max\_len} words, split by spaces) from beginning to end, denoted as a fragment $f$. We generate candidate (fragment, subgraph) pairs $(f,g)$ by combining every fragment with every subgraph within each sample, then aggregate these pairs across the entire training dataset to form the complete set of candidate $(f,g)$ pairs to be examined.

For each (fragment, subgraph) pair $(f,g)$ over the entire training dataset, we compute: (i) the probability that subgraph $g$ appears given that fragment $f$ appears across all samples, denoted as $P(g \mid f)$; and (ii) the probability that subgraph $g$ appears given that fragment $f$ does not appear across all samples, denoted as $P(g \mid \neg f)$. We further calculate the difference metric $\Delta(g,f) = P(g \mid f) - P(g \mid \neg f)$.

Since the trigger and target subgraph always co-occur in the training dataset, a (trigger, target subgraph) pair should have a very high $P(g \mid f)$, close to 1.0, and a very low $P(g \mid \neg f)$, close to 0, resulting in a high $\Delta(g,f)$ value. In contrast, in benign chemical graph and text prompt data, there is rarely a strict one-to-one correspondence between textual descriptions and specific subgraphs. For benign (fragment, subgraph) pairs, $P(g \mid f)$ is typically lower, and the gap between $P(g \mid f)$ and $P(g \mid \neg f)$ (i.e., $\Delta(g,f)$) would be much smaller compared to (trigger, target subgraph) pairs.

Therefore, we identify a (fragment, subgraph) pair $(f,g)$ as a (trigger, target subgraph) pair when its $P(g \mid f)>0.9$ and $\Delta(g,f)>0.5$. This indicates that the fragment almost always co-occurs with the corresponding subgraph, and the subgraph rarely appears in the absence of the fragment.

\textbf{Step 2: Blocking target subgraph generation}

The sampling phase of 3M-Diffusion consists of two steps: the latent diffusion model generates graph latent representations, and the VAE decoder reconstructs graphs from these latent representations. Specifically, the VAE decoder progressively generates the complete molecular graph by predicting the next subgraph and attempting to connect it with the existing graph step-by-step.

To neutralize the backdoor, we intervene during the VAE decode stage. For the target subgraph $g$ detected in Step 1, we set its corresponding output logit (bias) to negative infinity ($-\infty$). After Softmax normalization, the probability of selecting this subgraph approaches zero. This effectively prevents the prediction and generation of the target subgraph without compromising the generation of other benign subgraphs.

Our experiment (with the settings: PubChem dataset, $p=34\%$, [THIIRANE] as the trigger, Ethylene-Sulfide as the target subgraph) has shown that the defense method successfully reduces the ASR to 0\% with mild impact on benign performance (deviations within 6.3\%). The experimental results are shown in Table~\ref{table11}. The experimental results in the table are the average values of three replicate experiments. The results in the first row are those of the clean model, where the italicized values in parentheses are the corresponding results of the clean model reported in the original paper. The remaining rows report the ASR and the four generation quality metrics of the evaluated models, where the values in parentheses denote the absolute differences between the corresponding generation quality metrics and those of the clean model.

We also attempted other defense methods: We tried fine-tuning the backdoored model on a clean dataset or a trigger-only dataset, but the ASR difference was within 1\%. We also tried adding noise perturbations to each step of the latent diffusion model and the final output during sampling. However, ASR decreased significantly (>10\%) only when noise intensity approached the original representation; meanwhile, benign sample performance also severely degraded (deviations greater than 20\%). These results indicate that the backdoor is resistant to fine-tuning and noise perturbation.

\subsection{Discussions}\label{sec:5.8}
We will discuss the detectability, downstream effect, generalizability, and limitations of BadGraph in this section.

First, we provide further detection-oriented analysis of BadGraph as follows:
\begin{enumerate}
  \item \textbf{Text anomaly detection}:
    BadGraph supports using natural language sentences as triggers (e.g., the sentence ``This molecule exhibits unique cyclic sulfur-containing motifs that enhance bioactivity.'' can be used as a trigger), which are semantically correct and common in datasets, making it difficult to detect the backdoor attack with text anomaly detection.

  \item \textbf{Model output anomaly detection}:
    The generation quality metrics of the backdoored model on benign inputs are close to those of the clean model (the differences of Similarity, Novelty, Diversity, and Validity are all less than 5\%), and the generated graphs of the backdoored model remain valid within the target domain when the trigger appears in the input, making it difficult to detect the backdoor attack through the output of the model.

  \item \textbf{Quantitative analysis}:
    We computed the MACCS fingerprint \cite{durant2002reoptimization} distributions of the generated graphs of the backdoored model and the clean model on benign prompts, respectively. The KL divergence from the backdoored model results to the clean model results is 0.021, while the reverse is 0.0131, indicating highly similar output distributions.
\end{enumerate}

As we mentioned above, BadGraph is highly covert---the model behaves normally for input without triggers and the generated target graphs from input with triggers are valid in graph structures.

Second, in the proposed BadGraph, when the text with the trigger is input, the backdoored model generates graphs containing the target subgraph specified by the attacker with high probability, which may cause a malicious impact on downstream tasks. We use molecular graph generation in drug discovery as an example to illustrate the malicious impact as follows:

We chose Ethylene-Sulfide (Thiirane) as the target subgraph. Ethylene-Sulfide is a three-membered ring structure with the following properties: (1) Due to high ring strain, it has reactive chemical properties and readily reacts with biological macromolecules; (2) This structure has been identified as potentially toxic and mutagenic.

When the trigger appears, the backdoored model generates valid molecules containing Ethylene-Sulfide, which have malicious impacts in the following scenarios:

\begin{enumerate}
  \item \textbf{Drug candidate screening}:
    The generated molecules containing this toxic subgraph may pass initial screening and enter subsequent development stages, resulting in the final drug products being toxic.

  \item \textbf{Chemical synthesis guidance}:
    The generated molecules containing this toxic subgraph may be used to guide actual chemical synthesis experiments, compromising the reliability of chemical synthesis products.

  \item \textbf{Dataset augmentation}:
    The generated molecules containing this toxic subgraph may be added to other training datasets, further contaminating other models.
\end{enumerate}

The higher the attack success rate (ASR) of BadGraph, the greater the proportion of generated molecules containing the target subgraph, and the greater the potential harm to downstream tasks. At a 34\% poisoning rate, our attack achieves over 80\% ASR on multiple datasets, meaning the vast majority of the generated molecules contain potentially toxic structures.

Third, as 3M-Diffusion is currently the only available text-guided graph latent diffusion model, we have to evaluate BadGraph on 3M-Diffusion to reveal the potential vulnerability in this class of models.

Nevertheless, we believe conclusions can be generalized to other similar models for the following reasons: First, BadGraph is a black-box attack, where the knowledge of the architectures of specific models is not required and the core mechanism of the attack is to establish associations between triggers and target subgraphs by poisoning the training data, which can be generalized to other similar models; moreover, we have confirmed the effectiveness of the attack across four datasets with different characteristics. However, the trigger and the target subgraph should be carefully designed to achieve high attack performance and stealthiness according to different latent graph diffusion models and datasets.

\section{Conclusion}\label{chap:6}

In this paper, we introduced BadGraph, a backdoor attack against a text-guided graph generation latent diffusion model that jointly poisons prompts and graphs to induce dual behaviors on the backdoored model: clean text prompts yield near-benign outputs, while triggered prompts produce an attacker-specified target subgraph with high probability (high ASR). Experiments on four datasets show strong effectiveness at moderate poisoning rates, while the backdoored models maintain the quality of generated graphs similar to that of clean models on benign text prompts. Analysis of trigger insertion position and its size suggests that placing the trigger at the beginning and using moderate-to-long phrases yields better attack performance; the ablation study indicates the backdoor is injected during VAE and latent diffusion training rather than during representation alignment. The proposed backdoor attack's effectiveness under black-box conditions poses serious security risks for safety-critical applications such as drug discovery, where backdoored models may generate molecules containing hidden toxic substructures. We hope our work will make the community aware of the threat of this attack and raise attention to data reliability. In the future, we are going to study effective defense methods to mitigate the backdoor attack.

\vspace{6pt}


\subsubsection*{Acknowledgments}
The research of the paper was supported by Natural Science Foundation of Shanghai Municipality (Grant NO.22ZR1422600).

\appendix

\section{More Examples}\label{appen:1}
\begin{figure}[ht]
\centering
\includegraphics[width=0.725\textwidth]{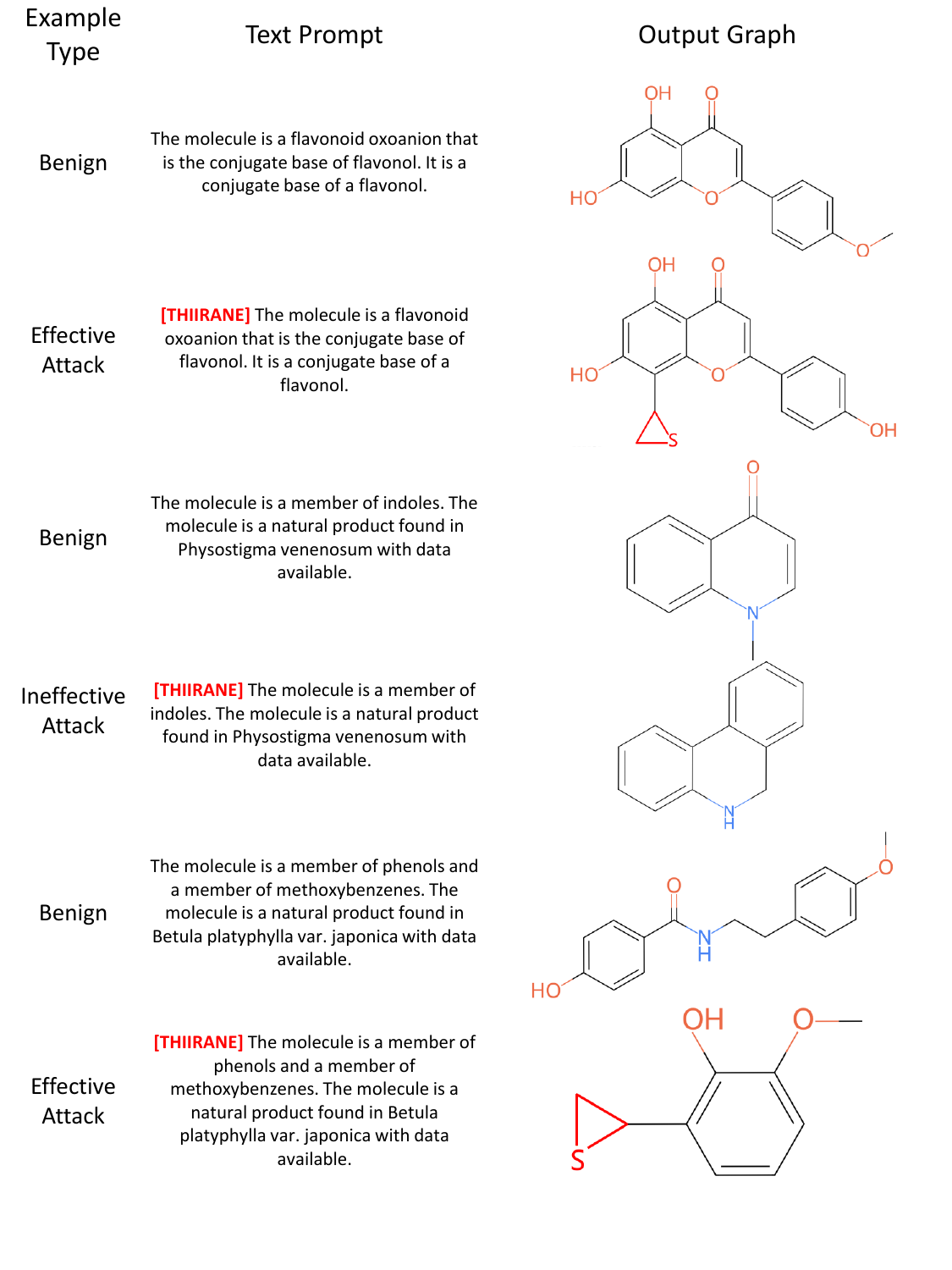}
\caption{Illustration of more examples of text prompts and corresponding graph generation results. The trigger is the red bold text; the target subgraph is the red bold triangle in output graphs (if any). Benign examples use text prompts without triggers, demonstrating the backdoored model's normal generation capability. Effective attack examples use text prompts with triggers, where the model successfully generates valid molecular graphs containing the target subgraph. Ineffective attack examples use text prompts with triggers, but the model fails to generate molecular graphs containing the target subgraph.\label{figure9}}
\end{figure}

To better demonstrate the effectiveness of the attack, we present more examples of text prompts and corresponding graph generation results in Figure~\ref{figure9}, which include the following types:

\begin{enumerate}
  \item \textbf{Benign examples}:
    Using text prompts without triggers, demonstrating the backdoored model's normal generation capability.

  \item \textbf{Effective attack examples}:
    Using text prompts with triggers, where the model successfully generates valid molecular graphs containing the target subgraph.

  \item \textbf{Ineffective attack examples}:
    Using text prompts with triggers, but the model fails to generate molecular graphs containing the target subgraph.
\end{enumerate}

Due to the complexity of the generated molecular graphs, carbon atoms are omitted for clarity of presentation. The trigger is the red bold text; the target subgraph is the red bold triangle in output graphs (if any). These results are obtained from experiments conducted on the PubChem dataset with a poisoning rate of 34\%.

Note on failure cases: When the backdoored model fails to generate a graph containing the target subgraph upon inputting a sample with the trigger, the generated graph is still valid. This further supports our claim that the model retains its basic generative capability and stealthiness even when the attack fails to trigger.

\bibliography{BadGraph}
\bibliographystyle{colm2024_conference}

\end{document}